\documentclass[journal]{IEEEtran}
\IEEEoverridecommandlockouts
\usepackage[T1]{fontenc}
\usepackage{balance}
\usepackage{silence}
\IEEEoverridecommandlockouts

\usepackage{cite}
\usepackage{graphicx}
\usepackage{picinpar}
\usepackage{stfloats}
\usepackage{tabularray}
\usepackage{url}
\usepackage{flushend}
\usepackage[latin1]{inputenc}
\usepackage{colortbl}
\usepackage{soul}
\usepackage{multirow}
\usepackage{pifont}
\usepackage{alltt}
\usepackage{enumerate}
\usepackage{siunitx}
\usepackage{pbox}
\usepackage{amsmath,amssymb,amsfonts}
\usepackage{amsthm}
\usepackage{algorithmic}
\usepackage{algorithm}
\usepackage{bm}
\usepackage{graphbox}
\usepackage{textcomp}
\usepackage{xcolor}
\usepackage{verbatim}
\usepackage[export]{adjustbox}
\usepackage{color}
\usepackage{tikz}
\def\BibTeX{{\rm B\kern-.05em{\sc i\kern-.025em b}\kern-.08em
    T\kern-.1667em\lower.7ex\hbox{E}\kern-.125emX}}

\definecolor{LightCyan}{rgb}{0.8,0.8,1.0}
\definecolor{LightRed}{rgb}{1.0,0.8,0.8}
\definecolor{LightGreen}{rgb}{0.8,1.0,0.8}
\definecolor{LightYellow}{rgb}{1.0,1.0,0.8}

\newcommand{\NA}[1]{$\spadesuit$\footnote{\color{red}{Nikolay: #1}}}

\DeclareMathOperator{\diag}{diag}

\newtheorem{example}{Example}

\ActivateWarningFilters[balance]

\makeatletter
\let\NAT@parse\undefined
\makeatother
\usepackage[hidelinks]{hyperref}

\newcommand{\calH}{{\cal H}}

\newcommand{\calN}{{\cal N}}

\newcommand{\calV}{{\cal V}}

\newcommand{\bfa}{\mathbf{a}}

\newcommand{\bff}{\mathbf{f}}
\newcommand{\bfg}{\mathbf{g}}
\newcommand{\bfh}{\mathbf{h}}

\newcommand{\bfs}{\mathbf{s}}

\newcommand{\bfu}{\mathbf{u}}

\newcommand{\bfx}{\mathbf{x}}

\newcommand{\bftheta}{{\boldsymbol{\theta}}}

\newcommand{\bfmu}{{\boldsymbol{\mu}}}

\newcommand{\bfpi}{{\boldsymbol{\pi}}}

\newcommand{\bfsigma}{{\boldsymbol{\sigma}}}

\newcommand{\bfomega}{{\boldsymbol{\omega}}}

\newcommand{\bfF}{\mathbf{F}}

\newcommand{\bfJ}{\mathbf{J}}
\newcommand{\bfK}{\mathbf{K}}
\newcommand{\bfL}{\mathbf{L}}
\newcommand{\bfM}{\mathbf{M}}

\newcommand{\bfQ}{\mathbf{Q}}
\newcommand{\bfR}{\mathbf{R}}

\newcommand{\bfV}{\mathbf{V}}

\newcommand{\bfX}{\mathbf{X}}

\newcommand{\bfZ}{\mathbf{Z}}

\newcommand{\bfPi}{\boldsymbol{\Pi}}

\title{Physics-Informed Multi-Agent Reinforcement \\ Learning  for Distributed Multi-Robot Problems}

\author{Eduardo Sebasti\'{a}n, \and Thai Duong, \and Nikolay Atanasov, \and Eduardo Montijano and Carlos Sag\"{u}\'{e}s%
\thanks{E. Sebasti\'{a}n, E. Montijano and C. Sag\"{u}\'{e}s are with the DIIS - I3A, Universidad de Zaragoza, Spain (e-mails: \texttt{\small \{esebastian, emonti, csagues\}@unizar.es}).}%
\thanks{T. Duong and N. Atanasov are with the Department of Electrical and Computer Engineering, University of California San Diego, La Jolla, CA 92093 USA (e-mails: \texttt{\small \{tduong, natanasov\}@ucsd.edu}).}%
\thanks{This work has been supported by ONR N00014-23-1-2353, N62909-24-1-2081 and NSF CCF-2402689 (ExpandAI), by Spanish projects PID2021-125514NB-I00, PID2021-124137OB-I00 and TED2021-130224B-I00 funded by MCIN/AEI/10.13039/501100011033, by ERDF A way of making Europe and by the European Union NextGenerationEU/PRTR, DGA T45-23R, a Spanish grant FPU19-05700 and a US-Spain Fulbright grant.}%
}

\newcommand\copyrighttext{%
  \footnotesize \textcopyright This paper has been accepted for publication at IEEE Transactions on Robotics. Please, when citing the paper, refer to the official manuscript with the following DOI: 10.1109/TRO.2025.3582836.}
\newcommand\copyrightnotice{%
\begin{tikzpicture}[remember picture,overlay]
\node[anchor=south,yshift=10pt] at (current page.south) {\fbox{\parbox{\dimexpr\textwidth-\fboxsep-\fboxrule\relax}{\copyrighttext}}};
\end{tikzpicture}%
}

\begin{document}
\maketitle
\copyrightnotice

\begin{abstract}
The networked nature of multi-robot systems presents challenges in the context of multi-agent reinforcement learning. 
Centralized control policies do not scale with increasing numbers of robots, whereas independent control policies do not exploit the information provided by other robots,
exhibiting poor performance in cooperative-competitive tasks. In this work we propose a physics-informed reinforcement learning approach able to learn distributed multi-robot control policies that are both scalable and make use of all the available information to each robot. Our approach has three key characteristics. First, it imposes a port-Hamiltonian structure on the policy representation, respecting energy conservation properties of physical robot systems and the networked nature of robot team interactions. 
Second, it uses self-attention to ensure a sparse policy representation able to handle time-varying information at each robot from the interaction graph. Third, we present a soft actor-critic reinforcement learning algorithm parameterized by our self-attention port-Hamiltonian control policy, which accounts for the correlation among robots during training while overcoming the need of value function factorization. Extensive simulations in different multi-robot scenarios demonstrate the success of the proposed approach, surpassing previous multi-robot reinforcement learning solutions in scalability, while achieving similar or superior performance (with averaged cumulative reward up to $\times2$ greater than the state-of-the-art with robot teams $\times6$ larger than the number of robots at training time). {We also validate our approach on multiple real robots in the Georgia Tech Robotarium under imperfect communication, demonstrating zero-shot sim-to-real transfer and scalability across number of robots.}
\end{abstract}



\begin{IEEEkeywords}
Cooperative control, distributed systems, multi-robot systems, physics-informed neural networks, reinforcement learning.
\end{IEEEkeywords}


\section{Introduction}\label{sec:intro}
\begin{figure}
    \centering
    \begin{tabular}{cc}
(a) navigation
&
(b) sampling
\\ 
\includegraphics[width=0.22\textwidth, height=0.20\textwidth]
{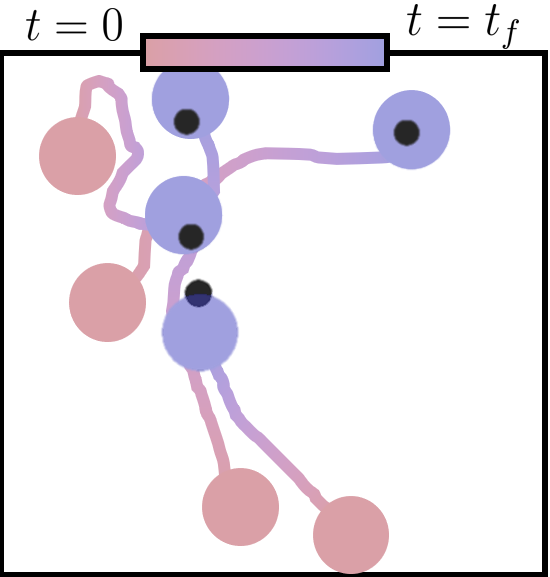} 
    & 
\includegraphics[width=0.22\textwidth, height=0.20\textwidth]{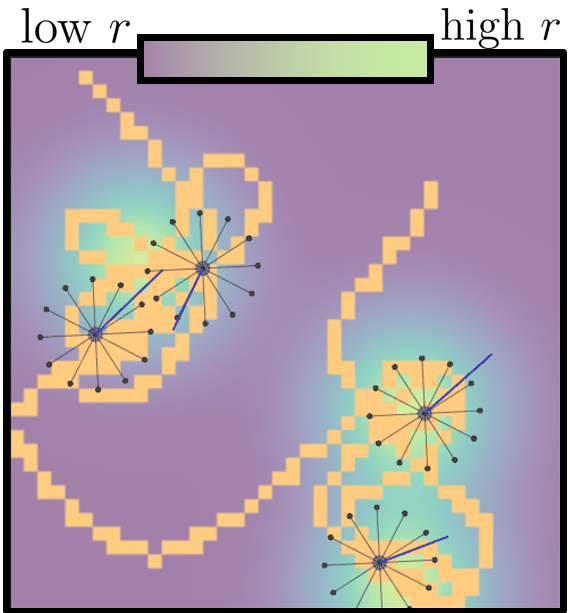}
\\
(c) transport
&
(d) grassland
\\
\includegraphics[width=0.22\textwidth, height=0.20\textwidth]
{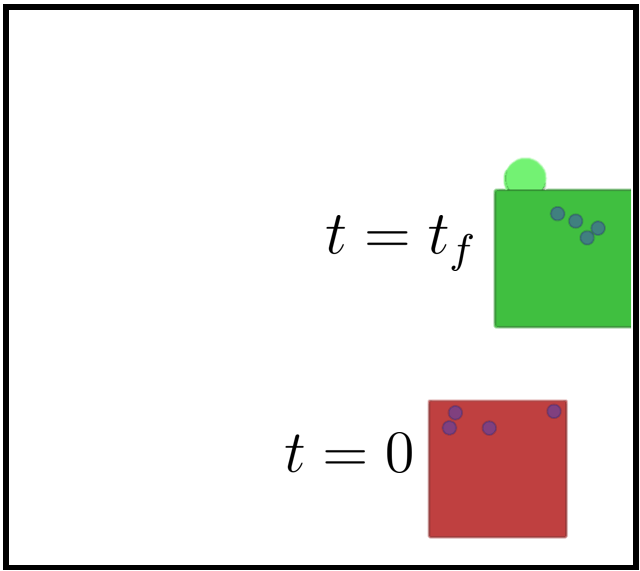} 
    & 
\includegraphics[width=0.22\textwidth, height=0.20\textwidth]{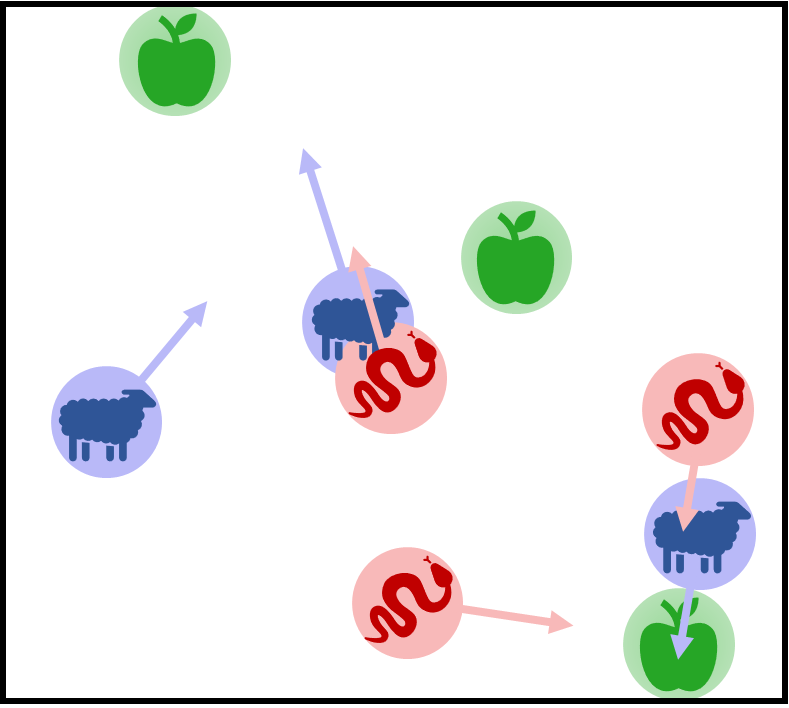}  
\\
\multicolumn{2}{c}{(e) {6x1-Half Cheetah}}
\\
\multicolumn{2}{c}{\includegraphics[width=0.30\textwidth] {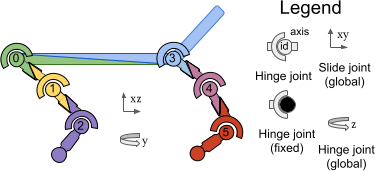}}
\\
\multicolumn{2}{c}{{(f) {multi-robot Robotarium \cite{pickem2017robotarium} navigation}}}
\\
\includegraphics[width=0.22\textwidth, height=0.15\textwidth]{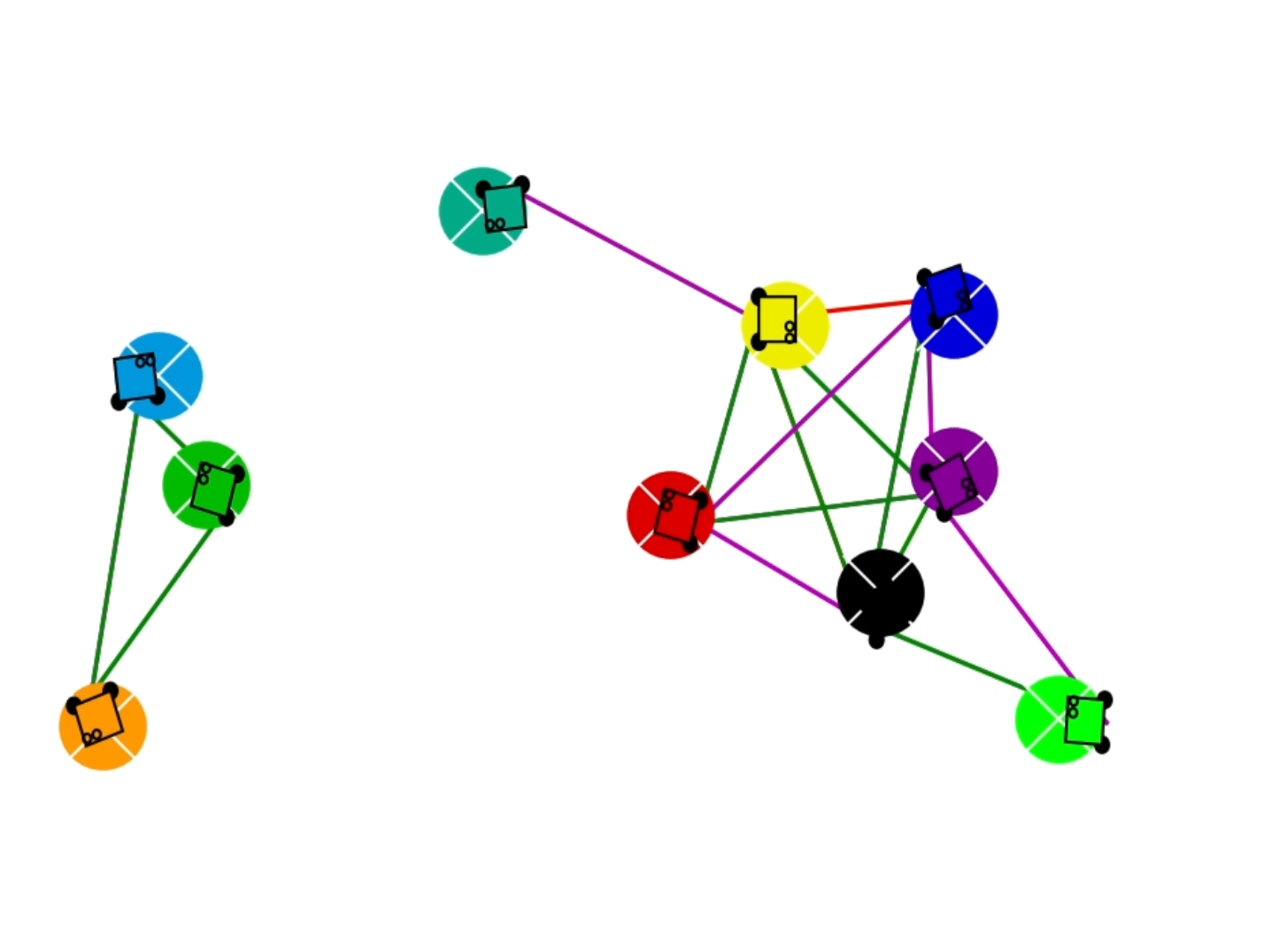} 
    & 
\includegraphics[width=0.22\textwidth, height=0.15\textwidth]{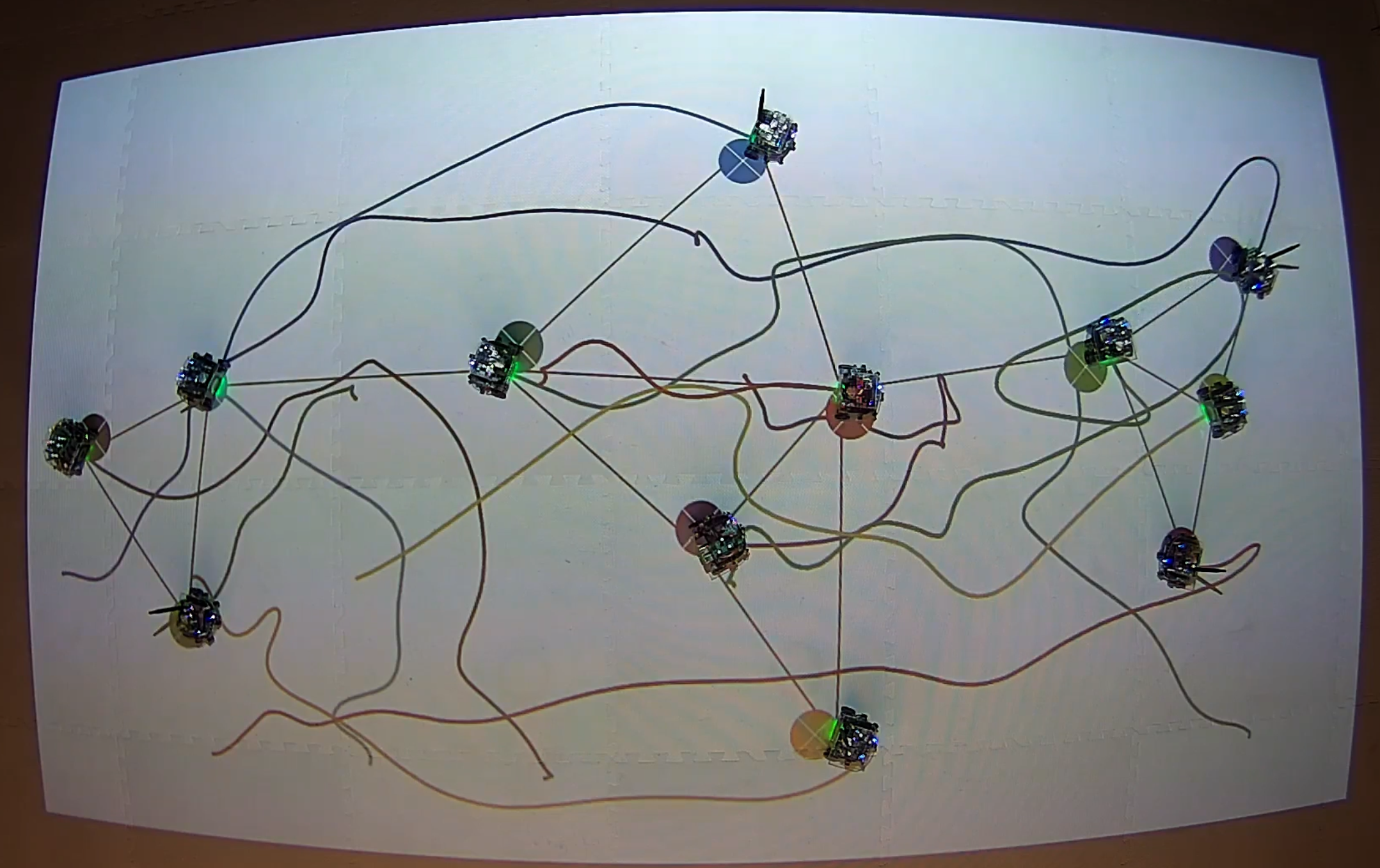} 
    \end{tabular}
    \caption{Examples of scenarios addressed by our physics-informed multi-agent reinforcement learning approach. The scenarios cover a wide variety of cooperative/competitive behaviors and levels of coordination complexity: {(a) robots learn to navigate to landmarks (black dots) while avoiding collisions {with other robots (colored trajectories from pink at $t=0$ to blue at $t=t_f$)}; (b) robots cooperate to do active LiDAR {(star pattern)} sampling {(orange areas refer to regions already sampled)} in an {unknown} environment with regions of {low reward (magenta areas) and high reward (green areas)}; (c) robots {(blue dots)} cooperate to transport a square box (red) {with unknown mass and that follows Newtonian dynamics} towards a desired region {(green dot, box turns green when it reaches the goal)}; (d) the robots (blue) collect food (green) while avoiding collisions with attackers (red), {since collision with attackers lead to deactivation of the robots}; (e) joints learn how to coordinate to make the {Half-Cheetah} robot walk, where each joint is an independent agent that can only interact with the adjacent agents (image from \cite{peng2021facmac}); {(f) a team of robots learns to navigate and avoid collisions in a real-world setting characterized by a tight space and high density of robots.}}}
\label{fig:overall_scenarios}
\end{figure}

\IEEEPARstart{M}{ulti-robot} systems promise improved efficiency and reliability compared to single robots in many applications, including exploration and mapping \cite{atanasov2015decentralized,tian2022kimera}, agriculture and herding \cite{kan2021task,pierson2015bio,sebastian2021multi,sebastian2022adaptive}, and search and rescue \cite{williams2021search}. However, the complexity of describing mathematically the objective and constraints in many of these problems makes the design of analytical controllers a challenging task.
Multi-agent reinforcement learning \cite{mataric1997reinforcement,matignon2007hysteretic, matignon2012coordinated,munikoti2023challenges,serra2023learning} addresses this issue by only requiring a high-level mathematical specification of the task (the reward function), which is commonly available.


A fundamental limitation of existing multi-agent reinforcement learning  approaches is the poor scalability they offer against increasing and time-varying numbers of robots.
Centralized control policies do not scale whereas independent control policies neglect the information that other robots can offer.
How to design and train control policies based on neural networks that are distributed and use all the available information is still an open problem~\cite{lo2023cheap,qu2022scalable,munikoti2023challenges}. 
The approach explored in this paper leverages physical knowledge about the robot system \cite{beckers2023learning,neary2023compositional, sebastian2023lemurs, nghiem2023physics, sanyal2023ramp, rodwell2023physics} to make this possible. Physics-informed neural networks are becoming popular in different fields, e.g., climate science, quantum mechanics or fluid dynamics \cite{cuomo2022scientific,xu2022physics}. They need less data for training and allow encoding general constraints found in physical systems. Nonetheless, they have not been used to learn distributed multi-robot control policies in multi-agent reinforcement learning.

The main contribution of our work is
a novel physics-informed multi-agent reinforcement learning approach suitable for general multi-robot problems (Sec.~\ref{sec:prosta}). The techniques to make this possible (Sec. \ref{sec:solution}) are summarized as follows.
\begin{itemize}
    \item We develop a novel distributed and scalable by design neural network architecture to describe multi-robot control policies. This is achieved by combining a physics-informed port-Hamiltonian description of the multi-robot system (Sec.~\ref{subsec:portHamiltonian}) with self-attention (Sec.~\ref{subsec:LEMURS}). The former naturally encodes the distributed nature of the policy and respects the energy conservation laws of the individual dynamics of the robots. The latter handles the information coming from communication or perception in time-varying neighborhoods. 
    
    \item We integrate the port-Hamiltonian self-attention policy in a soft actor-critic reinforcement learning algorithm that exploits the physics-informed description of the robot team to impose the sparsity pattern from the interaction graph in the policy function (Sec.~\ref{sec:sac}). To handle the networked nature of the multi-robot policy and avoid non-stationarity issues during training, we modify the acquisition of experience to keep track of the correlations among robots. 
\end{itemize}

Extensive simulations {and real experiments} in multi-robot scenarios (Sec.~\ref{sec:simulations}) demonstrate that the combination of multi-agent reinforcement learning techniques and a physics-informed description of the system achieves scalable control policies with similar or superior performance as the state-of-the-art (Sec.~\ref{sec:conclusion}).
    
{This paper is an extension of our prior conference paper \cite{sebastian2023lemurs}. The main difference of this work compared to \cite{sebastian2023lemurs} is that it addresses a multi-agent reinforcement learning problem (where only reward samples are provided), whereas \cite{sebastian2023lemurs} solves a learning-from-demonstrations problem (where data from an expert control policy is available). This leads to a new multi-agent reinforcement learning problem formulation and a substantially different approach beyond the policy parameterization, since our new approach involves modifications on the parameterization of the actor and the replay buffer of the actor-critic, besides some particularities in the reward function design and the critic. The neural networks that model the control policies are different as well, since the multi-agent reinforcement learning scenarios are continuous-time while those in \cite{sebastian2023lemurs} are discrete-time. Finally, all experiments in this paper are new, {including an extensive scalability analysis in a real multi-robot platform involving up to $16$ robots in a tight space}.}

\section{Related work}\label{sec:related}

\subsection{Learning multi-robot control policies from data}

The design of multi-robot control policies deals with two challenges. 

The first challenge is the mathematical formulation of the problem, where we must consider the task and the constraints inherited from the cooperative-competitive nature of the multi-robot team. To address this, recent works exploit machine learning and focus on learning control policies for optimal control or reinforcement learning problems \cite{bloembergen2015evolutionary,long2018towards, semnani2020multi}. When demonstrated data from an expert are available, inverse reinforcement learning \cite{ng2000algorithms} can be used to learn centralized \cite{dasari2020robonet,bogert2018multi} or distributed \cite{han2020cooperative, gharbi2023show} policies from task demonstrations. However, finding experts for multi-robot applications is difficult. It is also possible to apply supervised learning approaches to learn multi-robot control policies  \cite{zhu2021learning,zhou2019clone}, but, again, collecting labeled trajectories is hard. Therefore, in this work we use reinforcement learning, because a reward function is a high-level description of the task that is easy to build and is typically available. 

The second challenge is that the learning and execution of control policies for multi-robot systems should scale favorably with an increasing numbers of robots. Learning a joint policy function is challenging due to the exponential growth of the state and action space \cite{qu2020scalable}. Attention mechanisms are widely used in multi-agent reinforcement learning problems \cite{zambaldi2018deep, iqbal2019actor, li2020generative, parnika2021attention} to enhance the performance of centralized training settings where the agents are isolated from each other and do not consider communication during deployment. Graph neural networks \cite{munikoti2023challenges} have been utilized as a stable \cite{marino2023stability}, scalable and communication-aware policy representation in path planning, coverage, exploration, and flocking problems \cite{li2021message,khan2020graph, tolstaya2020learning,tolstaya2021multi,yang2021communication, gama2022synthesizing}.  These approaches assume discrete robot dynamics, fixed or known communication topology, or prior knowledge on the formulation of the control policy. In contrast, by using a port-Hamiltonian formulation and self-attention mechanisms, our approach directly learns control policies that handle time-varying neighbors, do not constrain the size of the neighborhoods, and learn constraints such as collision avoidance without specific mathematical formulation.

\subsection{Multi-agent reinforcement learning for robotic problems}

Multi-agent reinforcement learning  extends reinforcement learning approaches to problems where multiple agents interact in the environment \cite{kuyer2008multiagent,bucsoniu2010multi, vinyals2017starcraft, ellis2024smacv2, gronauer2022multi, oroojlooy2023review}. The first multi-agent reinforcement learning approaches considered centralized policies where the states, actions and observations of every agent are globally known by a central unit during deployment \cite{mataric1997reinforcement}. The complexity of these approaches exponentially scales with the number of agents, so they are not feasible in practice. In the absence of a centralized policy, the Markov game modeling the problem becomes non-stationary \cite{matignon2012independent,papoudakis2019dealing, bohmer2020deep}. The task is no longer stationary because the policies of all the other agents are changing, and so the environment itself. As a consequence, it is not possible to reach an equilibrium in training. 

To solve the non-stationarity issue, a common approach is to factorize the value and/or policy functions. The most common factorization, the so-called centralized-training decentralized-execution, departs from centralized-training reinforcement learning approaches \cite{haarnoja2018soft,lillicrap2015continuous,schulman2017proximal} and considers that the agents act independently. {To provide some global information to the agents, agents learn an approximation of the other agents' control policies~\cite{lowe2017multi, yu2022surprising, bettini2023benchmarl}. However, this proxy is only used during training and, since the resulting control policies only use the individual observations, the performance deteriorates the number of robots change. This aspect is partially overcome by considering heterogeneous agents \cite{kuba2021trust} but scalability is deteriorated again in large teams because the agents do not learn how to exploit neighboring information effectively.} 
Another option is to provide each agent with a global estimator that predicts the trajectories of the whole multi-robot team \cite{bloom2023decentralized} or to implicitly learn to coordinate with other agents through the value function \cite{bohmer2020deep}. In mean field approaches
\cite{yang2018mean}, the multi-agent game is reduced to an interaction between an agent and the average of the other agents, which is not practical. To overcome these issues, we propose a new formulation of the policy function that is distributed by design and captures the underlying interaction graph that describes the multi-robot system. 

Other works focus on more complex factorizations that account for local interactions among neighboring agents in the value function. For instance, the value function can be approximated as dependent on the neighbors only \cite{wang2022darl1n}. {Either using analytical \cite{qu2020scalable, qu2022scalable, de2020independent} or learned \cite{motokawa2023interpretability,kortvelesy2022qgnn, hu2023graph} factorizations, the control policies are still restricted to the same team size used during training because the policies do not explicitly consider communication. Closely related to ours, the authors of \cite{huang2024collision} propose an end-to-end neural network based on attention to achieve scalability of the control policy in quadrotor swarm navigation problems. Similarly,} by using a port-Hamiltonian description of the system, our approach avoids factorizing the reward or the value functions, thus fully exploiting the available information in the experience while learning scalable distributed control policies.


\subsection{Physics-informed neural networks for robotic problems}

While black-box neural networks are widely used for learning control policies, they do not encode energy conservation and kinematic constraints satisfied by physical robot systems. Failing to infer them from data may result in unstable behaviors. Moreover, other constraints are present that come from the perception and communication modules of the robots and which restrict the information available to the robot. Similar issues are found in other physical applications \cite{xu2022physics}, leading to physics-informed neural network \cite{cuomo2022scientific}, neural networks that use the differential equations that model physical systems as building blocks. The use of physics-informed machine learning for robotics and control is very recent \cite{nghiem2023physics,sanyal2023ramp,rodwell2023physics,zhao2021physics,sartoretti2019distributed} and focuses on centralized controllers. Nevertheless, physics-informed neural network can also be used to address the learning of cooperative distributed control policies.

Many dynamical systems--from robots interacting with their surroundings to large-scale multi-physics systems--involve a number of interacting subsystems. This compositional property can be exploited \cite{neary2023compositional} to train neural network sub-models from data generated by simple sub-systems, and the dynamics of more complex composite systems are then predicted without requiring additional data. The systems are represented as a port-Hamiltonian neural network \cite{beckers2023learning}, a class of neural ordinary differential equations that uses a port-Hamiltonian dynamics formulation as inductive bias \cite{van2014port}. A key contribution of our work is to represent the robot team as a port-Hamiltonian system and learn distributed control policies by modeling robot interactions as energy exchanges. The use of Hamiltonian mechanics has been explored for centralized control policies or distributed but fixed-time known topologies~\cite{furieri2021distributed, galimberti2023hamiltonian,shi2020neural}. Meanwhile, our work achieves scalability with a time-varying topology by modeling robot interactions using self-attention~\cite{vaswani2017attention}.


\section{Problem formulation}\label{sec:prosta}
Consider a team of robots, indexed by $\mathcal{V} = \{1, \hdots, n \}$. The robot team motion is governed by \textit{known} continuous-time control-affine stochastic dynamics:
\begin{equation}\label{eq:affinedynamics}
    \dot{\mathbf{x}}(t) = \mathbf{f}(\mathbf{x}(t),\mathbf{u}(t)) + \mathbf{L}\bf{\omega}(t),
\end{equation}
where $\mathbf{x}(t)=[(\mathbf{x}^1(t))^{\top}, \hdots, (\mathbf{x}^n(t))^{\top}] \in \mathcal{X} \subseteq \mathbb{R}^{n \times n_x}$ is the joint state of the robot team at time $t \geq 0$, with $\mathbf{x}^i(t)$ the state of robot $i$ at time $t$. On the other hand, \mbox{$\mathbf{u}=[(\mathbf{u}^1(t))^{\top}, \hdots, (\mathbf{u}^n(t))^{\top}] \in \mathcal{U} \subseteq \mathbb{R}^{n \times n_u}$} is the joint input, with $\mathbf{u}^i(t)$ the input of robot $i$ at time $t$. The term $\bfomega(t)$ is white noise modeling the uncertainty in the robot sensors and actuators and $\Xi = \mathbf{L}\mathbf{L}^\top$ is the noise diffusion matrix. Let $\{t_{\tau}\}_{\tau=0}^{\infty}$ be a sequence of discrete time instants such that $t_{\tau+1} - t_{\tau} = T_{\tau} > 0$ and assume zero-order hold inputs such that $\mathbf{u}(t) = \mathbf{u}(t_\tau)$, $\forall t \in [t_\tau, t_{\tau+1})$. Then, from Eq.~\eqref{eq:affinedynamics}, we can obtain an Euler discretization of the dynamics with discrete-time state $\mathbf{s}_\tau = \mathbf{x}(t_\tau)$, action $\mathbf{a}_\tau = \mathbf{u}(t_\tau)$, and dynamics:
\begin{equation}\label{eq:discretized}
    \mathbf{s}_{\tau+1} = \mathbf{s}_\tau + T_{\tau} \mathbf{f}(\mathbf{s}_\tau,\mathbf{a}_\tau) + \mathbf{n}_\tau,
\end{equation}
where $\mathbf{n}_\tau$ is zero-mean Gaussian noise with covariance $T \Xi$. We will formulate the dynamics $\mathbf{f}$ using port-Hamiltonian mechanics as its modularity in terms of energy effectively describes the networked interactions in a robot team.

The multi-robot task and the interactions among the robots are modeled as a Markov Decision Process (MDP), defined as a tuple $(\mathcal{X}, \mathcal{U}, p, r, \gamma)$. In the tuple, $p: \mathcal{X} \times \mathcal{X} \times \mathcal{U} \rightarrow \mathbb{R}$ is the probability density of the next joint state $\bfs_{\tau+1}$ conditioned on the current joint state $\mathbf{s}_\tau$ and joint action $\bfa_\tau$, \mbox{$r: \mathcal{X} \times \mathcal{U} \rightarrow [r_{\min}, r_{\max} ]$} is a reward function encoding the objectives of the multi-robot task, and $\gamma \in (0, 1)$ is the discount factor. Whereas the robot dynamics are formulated in continuous time, the MDP is formulated in discrete time based on the zero-order hold discretization in Eq.~\eqref{eq:discretized}. Accordingly, $p(\mathbf{s}_{\tau+1}|\mathbf{s}_{\tau}, \mathbf{a}_{\tau})$ is a Gaussian density with mean $\mathbf{s}_{\tau} + T \bff(\mathbf{s}_{\tau}, \mathbf{a}_{\tau})$ and covariance $T \Xi$.



The robots interact in a distributed manner, described by a time-varying undirected graph $\mathcal{G}_\tau = (\mathcal{V},\mathcal{E}_\tau)$, where $\mathcal{E}_\tau \subseteq \mathcal{V} \times \mathcal{V}$ is the set of edges. An edge $(i,j) \in \mathcal{E}_\tau$ exists when robots $i$ and $j$ interact at time $\tau$. Let $\mathbf{A}_\tau \in \{0,1\}^{n \times n}$ be the adjacency matrix associated to $\mathcal{G}_\tau$, such that $[\mathbf{A}_\tau]_{ij} = 1$ if and only if $(i,j) \in \mathcal{E}_\tau$, and $0$ otherwise. The set of $k$-hop neighbors of robot $i$ at time $\tau$ is $\mathcal{N}^{i,k}_\tau = \{j \in \mathcal{V} \mid [\mathbf{A}^k_\tau]_{ij} \neq 0 \}$, {where $\mathbf{A}^k_\tau$ is the $k$-th power of matrix $\mathbf{A}_{\tau}$}. We remark that $\mathcal{N}^{i,k}_\tau$ includes robot $i$.

The goal of this work is to learn distributed control policies that solve a given multi-robot task, such that they respect the networked structure of the multi-robot team and the MDP model. We represent the policies as stochastic Markov control policies that depend on the $k$-hop neighbors of robot $i$:
\begin{equation}\label{eq:general_controller}
    \mathbf{a}^i_\tau \sim \bfpi_{\bftheta}\left(\mathbf{a}^i_\tau |\bfs^i_{\calN_\tau^{i,k}}\right).
\end{equation}
Here, $\bfs^i_{\calN_\tau^{i,k}} = \{\bfs_\tau^j | j \in \calN_\tau^{i,k}\}$ denotes the states of robot $i$ and its $k$-hop neighbors, and $\bftheta$ denotes the control policy parameters. The use of a stochastic Markov policy is not only motivated by the MDP model but also by the fact that distributed control policies are prone to uncertainty from the communication-perception modules, control goals, and interaction with the environment.
We assume that the policy in Eq.~\eqref{eq:general_controller} is the same for all the robots. 
The joint control policy of all the robots is denoted by \mbox{$\bfPi_{\bftheta} = [\bfpi_{\bftheta}^{\top}\left(\mathbf{a}^1_\tau |\bfs^1_{\calN_\tau^{1,k}}\right), \hdots, \bfpi_{\bftheta}^{\top}\left(\mathbf{a}^n_\tau |\bfs^n_{\calN_\tau^{n,k}}\right)]^{\top}$}.

Learning a distributed control policy $\bfpi_\bftheta$ that solves a certain multi-robot task is equivalently posed as learning the parameters $\bftheta$ such that $\bfpi_{\bftheta}$ maximizes the expected sum of rewards over time, i.e.,
\begin{align}\label{eq:max_action_value_function}
    &\max_{\bftheta} \text{ } Q_{\bfPi_{\bftheta}}(\bfs, \bfa) =
    \\
    &\max_{\bftheta}  \text{ }\mathbb{E}_{\mathbf{s}_\tau \sim p}\mathbb{E}_{\mathbf{a}_\tau \sim \bfPi_{\bftheta}(\cdot|\mathbf{s}_\tau)}\left[\sum_{\tau=0}^{\infty}\gamma^\tau r(\mathbf{s}_\tau, \mathbf{a}_\tau)|\mathbf{s}_0 = \mathbf{s}, \mathbf{a}_0 = \mathbf{a}\right].  \notag  
\end{align}
In Eq.~\eqref{eq:max_action_value_function}, the function $Q_{\bfPi_{\bftheta}}(\bfs, \bfa)$ is known as the action-value ($Q$) function associated with the policy $\bfPi_{\bftheta}$ \cite{bucsoniu2010multi}.

We do not make any assumptions on the reward function or action-value function, such as a specific factorization \cite{qu2020scalable, qu2022scalable}. 
The purpose of this work is to design a control policy that maximizes $Q_{\bfPi_{\bftheta}}(\bfs, \bfa)$ and enforces the distributed factorization expressed in Eq.~\eqref{eq:general_controller}.


\section[Physics-informed multi-agent reinforcement learning]{Physics-informed multi-agent \\ reinforcement learning}\label{sec:solution}

In this section, we present a novel physics-informed multi-agent reinforcement learning approach to find distributed control policies that solve multi-robot tasks 
as defined in Eq.~\eqref{eq:max_action_value_function}, under the restrictions on the robot dynamics and available information defined in Eq. \eqref{eq:general_controller}. Our formulation is done in continuous time, following the continuous-time definition of the robot dynamics in Eq.~\eqref{eq:affinedynamics}. Later, zero-order hold control is used to utilize a discrete-time version of the control policies in the reinforcement learning algorithm.

First, we present a port-Hamiltonian formulation of the multi-robot dynamics and an energy-based distributed control design that can shape the interactions and the Hamiltonian of the closed-loop system (Sec.~\ref{subsec:portHamiltonian}). Given experience from trial and error simulations, we employ a self-attention mechanism to learn the parameters of the control policy that maximize the cumulative reward (Sec. \ref{subsec:LEMURS}). 
To simplify the notation, we omit the time dependence of the states $\bfx$ and controls $\bfu$ in the reminder of the paper. To facilitate the exposition, in Sec. \ref{subsec:portHamiltonian} and \ref{subsec:LEMURS} we assume that the robot dynamics and control policies are deterministic. In Sec. \ref{sec:sac}, we explain how to return to the stochastic setting.

\subsection{Port-Hamiltonian dynamics for multi-robot energy conservation}\label{subsec:portHamiltonian}

Port-Hamiltonian mechanics are a general yet interpretable modeling approach for learning and control. On the one hand, many physical networked systems can be described as a port-Hamiltonian system~\cite{furieri2021distributed} using the same formulation and with a modular and distributed interpretation. Meanwhile, the port-Hamiltonian description allows to derive general energy-based controllers with closed-loop stability guarantees. Since robots are physical systems that satisfy Hamiltonian mechanics, we model each robot as a port-Hamiltonian system \cite{van2014port}:
\begin{equation}\label{eq:open_loop_intro_individual}
        \dot{\bfx}^i = \left(\mathbf{J}^{i}(\bfx^i) - \mathbf{R}^{i}(\bfx^i)\right) \frac{\partial {H}^i(\bfx^i)}{\partial\bfx^i}
        + \mathbf{F}^i(\bfx^i)\bfu^i,
\end{equation}
where the skew-symmetric interconnection matrix $\mathbf{J}^i(\bfx^i)$ represents energy exchange within a robot, the positive-semidefinite dissipation matrix $\bfR^i(\bfx^i)$ represents energy dissipation, the Hamiltonian ${H}^i(\bfx^i)$ represents the total energy, and the matrix $\bfF^i(\bfx_i)$ is the input gain.

The interconnection of port-Hamiltonian systems leads to another port-Hamiltonian system \cite{shaft2004port}. Therefore, if the control and state of each robot are considered as input and output energy ports, then, due to the modularity of port-Hamiltonian dynamics, the multi-robot system with joint state $\bfx$ also follows port-Hamiltonian dynamics:
\begin{equation}\label{eq:open_loop_intro}
        \dot{\bfx} =  \left(\mathbf{J}(\bfx)  -  \mathbf{R}(\bfx)\right)\frac{\partial {H}(\bfx)}{\partial\bfx}  +  \mathbf{F}(\bfx)\bfu,
\end{equation}
where $\mathbf{J}(\bfx)$, $\bfR(\bfx)$, and $\bfF(\bfx)$ are block-diagonal:
\begin{equation}\label{eq:block_diag_structure}
\begin{aligned}
    \bfJ(\bfx) &= \text{diag}\left(\bfJ^{1}(\bfx^1), \ldots, \bfJ^{n}(\bfx^n) \right), \\
    \bfR(\bfx) &= \text{diag}\left(\bfR^{1}(\bfx^1), \ldots, \bfR^{n}(\bfx^n) \right), \\
    \bfF(\bfx) &= \text{diag}\left(\bfF^{1}(\bfx^1), \ldots, \bfF^{n}(\bfx^n) \right), \\
\end{aligned}
\end{equation}
and $H(\bfx) = \sum_{i=1}^n H^{i}(\bfx^i)$. It is noteworthy that the expression in Eq.~\eqref{eq:open_loop_intro} is control-affine and follows the definition of Eq.~\eqref{eq:affinedynamics}, with $\bff(\bfx, \bfu) = \bfh(\bfx) + \bfg(\bfx)\bfu$, \mbox{$\bfh(\bfx) = \left(\mathbf{J}(\bfx)  -  \mathbf{R}(\bfx)\right)\frac{\partial {H}(\bfx)}{\partial\bfx}$}, $\bfg(\bfx) = \bfF(\bfx)$, and $\bfL = \bf0$ since we are considering a deterministic setting for now.

Without control, the trajectories of the open-loop system in \eqref{eq:open_loop_intro} would follow the dynamics of the robots in the absence of interactions with the environment or other robots. The dynamics need to be controlled by a policy 
in order to accomplish the desired task. We propose to design a control policy $\bfmu_{\bftheta}(\bfx)$ and, then, obtain the desired policy in Eq.~\eqref{eq:general_controller}. Policy $\bfmu_{\bftheta}(\bfx)$ is designed using an interconnection and damping assignment passivity-based control (IDA-PBC) approach \cite{van2014port}, which injects additional energy in the system through the control input $\bfu$ to obtain closed-loop dynamics that achieve the desired task:
\begin{equation}\label{eq:closed_loop_intro}
        \dot{\bfx} = \left( \mathbf{J}_{\bftheta}(\bfx)   -   \mathbf{R}_{\bftheta}(\bfx)\right)
\frac{\partial {H_{\bftheta}}(\bfx)}{\partial\bfx}, 
\end{equation}
with Hamiltonian $H_{\bftheta}(\bfx)$, skew-symmetric interconnection matrix $\mathbf{J}_{\bftheta}(\bfx)$, and positive semidefinite dissipation matrix $\mathbf{R}_{\bftheta}(\bfx)$, which depend on the control policy $\bfmu_{\bftheta}(\bfx)$. By matching the terms in \eqref{eq:open_loop_intro} and \eqref{eq:closed_loop_intro}, one obtains the joint policy:
\begin{align}\label{eq:open_loop_intro_controller}
\bfu  &= \bfPi_{\bftheta}(\bfx) = \\
& \!\!\mathbf{F}^{\dagger}(\bfx)\!
\left( \left(\mathbf{J}_{\bftheta}(\bfx) \!-\! \mathbf{R}_{\bftheta}(\bfx)\right)\frac{\partial {H_{\bftheta}}(\bfx)}{\partial\bfx} \!-\! \left(\mathbf{J}(\bfx) \!-\! \mathbf{R}(\bfx)\right)\frac{\partial {H}(\bfx)}{\partial\bfx} \right),\notag
\end{align}
where $\mathbf{F}^{\dagger}(\mathbf{x}) = \left(\bfF^{\top}(\bfx)\bfF(\bfx)\right)^{-1}\bfF^{\top}(\bfx)$ is the pseudo-inverse of $\mathbf{F}(\bfx)$. 

If the robots are fully-actuated, i.e., $\bfF(\bfx)$ is full-rank,
then the input $\bfu$ in \eqref{eq:open_loop_intro_controller} exactly transforms the open loop system in \eqref{eq:open_loop_intro} to the closed-loop system in \eqref{eq:closed_loop_intro}. For underactuated systems, the transformation may not be exact \cite{blankenstein2002matching}. Being able to maximize $Q_{\bfPi_{\bftheta}}$ is, hence, related to whether the robot configurations that solve the task are realizable by the class of control policies in \eqref{eq:open_loop_intro_controller}. Even if goals of the task are not realizable, the policy parameters $\bftheta$ may still be optimized to achieve a behavior as good as possible to solve the task.

Let $[\mathbf{J}_{\bftheta}(\bfx)]_{ij}$ and $[\mathbf{R}_{\bftheta}(\bfx)]_{ij}$ denote the $n_x \times n_x$ blocks with index $(i,j)$, representing the energy exchange between robot $i$ and $j$ and the energy dissipation of robot $i$ caused by robot $j$, respectively. Since the input gain $\mathbf{F}(\bfx)$ in \eqref{eq:block_diag_structure} is block-diagonal, the individual control policy of robot $i$ is:
\begin{align}\label{eq:open_loop_intro_controller_ind}
\kern -5pt \bfmu_{\bftheta}(\bfx) &= (\mathbf{F}^{i})^{\dagger}(\bfx^i)
\biggl(  \sum_{j \in \calV}\left([\mathbf{J}_{\bftheta}(\bfx)]_{ij} - [\mathbf{R}_{\bftheta}(\bfx)]_{ij}\right)\frac{\partial H_{\bftheta}(\bfx)}{\partial \bfx_j} \notag\\
& \qquad \quad \kern 0.3cm- \left(\mathbf{J}^{i}(\bfx^i) - \mathbf{R}^i(\bfx^i)\right)\frac{\partial {H}^i(\bfx)}{\partial\bfx^i} \biggr).
\end{align}
Note that the individual control policy $\bfmu_{\bftheta}(\bfx)$ in \eqref{eq:open_loop_intro_controller_ind} does not necessarily respect the hops in the communication network as desired in \eqref{eq:general_controller} because it depends on the structure of $\mathbf{J}_{\bftheta}(\bfx)$,  $\mathbf{R}_{\bftheta}(\bfx)$, and $H_{\bftheta}(\bfx)$. In Sec.~\ref{subsec:LEMURS}, we impose conditions on these terms to ensure that they respect the communication topology and are skew-symmetric, positive semidefinite and positive respectively, as required for a valid port-Hamiltonian system and to find a policy $\bfmu_{\bftheta}$ that only depends on the $k$-hop neighbors.

\subsection{Self-attention parameterization to enforce communication patterns}\label{subsec:LEMURS}

\begin{figure*}[t]
    \centering
    \includegraphics[width=\linewidth, trim={0 12cm 0 0}, clip]{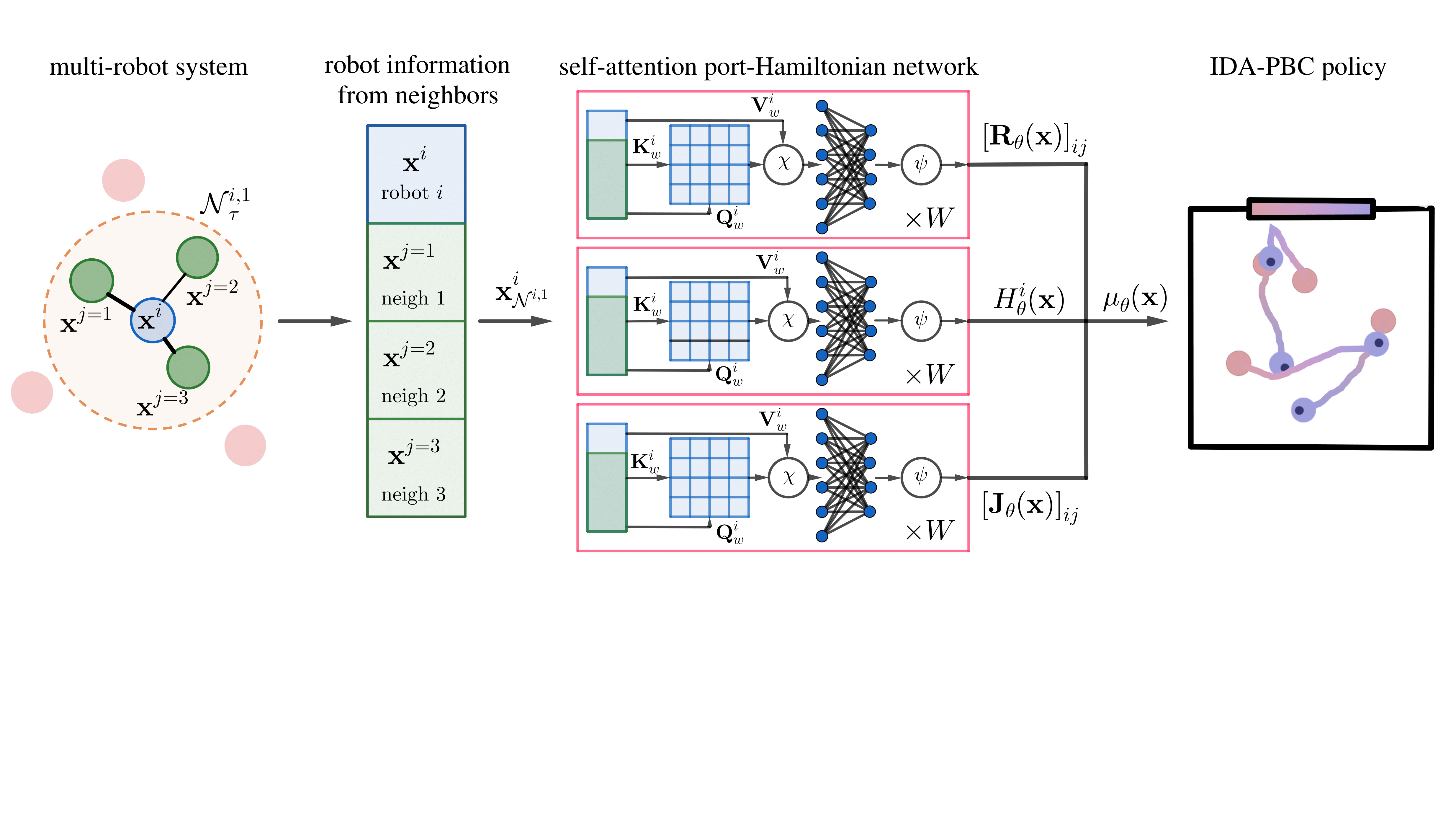}
    \caption{\small{{Physics-informed policy parameterization. At each instant, robot $i$ receives state information from its neighbors, typically associated to a perception or communication range. Then, three self-attention-based modules, each one associated to a component of the desired closed-loop port-Hamiltonian dynamics ($[\mathbf{R}_{\theta}(\mathbf{x})]_{ij}$, $[\mathbf{J}_{\theta}(\mathbf{x})]_{ij}$, $H_{\theta}^i(\mathbf{x}$)), use the information from the neighbors to compute the parameters of an interconnection and damping assignment passivity-based control (IDA-PBC) policy, which is then used to execute the desired control action.}}}
    \label{fig:architecture}
\end{figure*}

We seek to learn distributed control policies that follow the structure of Eqs.~\eqref{eq:open_loop_intro_controller}-\eqref{eq:open_loop_intro_controller_ind} and (i) scale with the number of robots, (ii) handle time-varying communication and (iii) guarantee the port-Hamiltonian constraints. To do so, we first derive conditions on the port-Hamiltonian terms of the controller, $\mathbf{J}_{\bftheta}(\mathbf{x})$, $\mathbf{R}_{\bftheta}(\mathbf{x})$ and ${H}_{\bftheta}(\mathbf{x})$, which are the terms to be learned from the reinforcement learning experience. Then, we develop a novel architecture based on self-attention to ensure that the learned control policies guarantee the desired requirements. We summarize the overall neural network architecture in Fig.~\ref{fig:architecture}.

To respect the robot team topology defined by the graph $\mathcal{G}$, we first impose $\mathbf{J}_{\bftheta}(\mathbf{x})$ and $\mathbf{R}_{\bftheta}(\mathbf{x})$ to be block-sparse, 
\begin{equation} 
\label{eq:JRH_conditions1}
        [\mathbf{J}_{\bftheta}(\mathbf{x})]_{ij} =
        [\mathbf{R}_{\bftheta}(\mathbf{x})]_{ij} = \mathbf{0}, \quad \forall j \notin \calN^{i,k}.
\end{equation}
From the perspective of robot $i$, this means that the controller only considers information from its $k$-hop neighbors. Moreover, we require that the desired Hamiltonian factorizes over the $k$-hop neighborhoods:
\begin{equation} \label{eq:individual_Hi}
        H_{\bftheta}(\mathbf{x}) = \sum_{i=0}^n H_{\bftheta}^{i}(\bfx^i_{\calN^{i,k}}),
\end{equation}
with $\bfx^i_{\calN^{i,k}} = \{\bfx^j | j \in \calN^{i,k}\}$. The factorization in \eqref{eq:individual_Hi} ensures that each robot $i$ can calculate
\begin{equation}\label{eq:partial_partial}
    \frac{\partial H_{\bftheta}(\bfx)}{\partial \mathbf{x}^i} = \sum_{j\in \calN^{i,k}}\frac{\partial H^{j}_{\bftheta}(\bfx^j_{\calN^{j,k}})}{\partial \mathbf{x}^i}
\end{equation}
by gathering $\partial H^{j}_{\bftheta}(\bfx^j_{\calN^{j,k}})/\partial \mathbf{x}^i$ from its $k$-hop neighbors. Then, the control policy $\bfmu_{\bftheta}$ of robot $i$ becomes:
\begin{align}
\bfmu_{\bftheta}(\bfx)  &=  (\mathbf{F}^{i}(\mathbf{x}^i))^\dagger
\kern -2pt \biggl( \sum_{j \in \calN^{i,k}}\kern -0.2cm \left([\mathbf{J}_{\bftheta}(\mathbf{x})]_{ij} \kern -0.1cm-\kern -0.1cm [\mathbf{R}_{\bftheta}(\mathbf{x})]_{ij}\right)\frac{\partial H_{\bftheta}(\mathbf{x})}{\partial \mathbf{x}^j}  
\notag\\
& \qquad \qquad  -\left(\mathbf{J}^{i}(\mathbf{x}^i) - \mathbf{R}^{i}(\mathbf{x}^i)\right)\frac{\partial {H}^{i}(\mathbf{x})}{\partial\mathbf{x}^i} \biggr). \label{eq:open_loop_intro_controller_ind_distributed}
\end{align}

Imposing the requirements in \eqref{eq:JRH_conditions1}-\eqref{eq:individual_Hi} is a first step towards making the control policy in \eqref{eq:open_loop_intro_controller_ind_distributed} distributed. Note that the terms $[\mathbf{J}_{\bftheta}(\mathbf{x})]_{ij}$ and $[\mathbf{R}_{\bftheta}(\mathbf{x})]_{ij}$ might still depend on the joint state vector $\bfx$ even though the sum runs over the $k$-hop neighbors in $\calN^{i,k}$. Next, we discuss how to remove this dependence and achieve a similar factorization as \eqref{eq:individual_Hi}.

First, we model $[\mathbf{J}_{\bftheta}(\bfx)]_{ij}, [\mathbf{R}_{\bftheta}(\bfx)]_{ij}$, and $H_{\bftheta}^{i}(\bfx)$ in
Eq.~\eqref{eq:open_loop_intro_controller_ind_distributed} with the parameters $\bftheta$ shared across the robots, so that the team can handle time-varying communication graphs.
Specifically, we propose a novel architecture 
based on self-attention~\cite{vaswani2017attention}. Self-attention layers extract the relationships among the inputs of a sequence by calculating the importance associated to each input using an attention map. The length of the sequences can vary as the number of parameters of the self-attention is constant with the number of inputs. Our key idea is to consider the self and neighboring states as the sequence, where each neighbor's state is an input. We now detail how to model each of the port-Hamiltonian terms. 

To learn
$[\mathbf{R}_{\bftheta}(\bfx)]_{ij},$
robot $i$ will use, at instant $t$, the state $\bfx^j$ from all $k$-hop neighbors $j \in \mathcal{N}^{i,k}$. The sequence of states is given by $\bfx^i_{\calN^{i,k}} $.
The proposed architecture is composed by a sequence of layers, indexed using the subscript $w=1,\hdots, W$. Following a self-attention mechanism, each layer computes the operations presented below:
\begin{gather}\begin{aligned}\label{eq:R_layers}
        &\kern -0.2cm \mathbf{Q}^i_1 =  \mathbf{A}_1^{\mathbf{R}}\bfx^i_{\calN^{i,k}}, \hbox{ } \mathbf{K}^i_1 = \mathbf{B}_1^{\mathbf{R} }\bfx^i_{\calN^{i,k}}, \hbox{ } \mathbf{V}^i_1 = \mathbf{C}_1^{\mathbf{R}}\bfx^i_{\calN^{i,k}}
    \\
    &\kern -0.2cm\mathbf{Y}^i_1 = \chi\left(\mathsf{softmax}\left(\frac{\beta(\mathbf{Q}^i_1)\beta((\mathbf{K}^i_1)^{\top})}{\sqrt{|\mathcal{N}^{i,k}|}}\right) \beta(\mathbf{V}^i_1)\right),
    \\
    &\kern -0.2cm\mathbf{X}^i_1 = \psi(\mathbf{D}_1^{\mathbf{R}}{\mathbf{Y}}_1^i),
    \\
    & \kern -0.2cm \vdots
    \\
    &\kern -0.2cm \mathbf{Q}^i_w =  \mathbf{A}_w^{\mathbf{R}}{\mathbf{X}}^i_{w-1}, \hbox{ } \mathbf{K}^i_w = \mathbf{B}_w^{\mathbf{R} }{\mathbf{X}}^i_{w-1}, \hbox{ } \mathbf{V}^i_w = \mathbf{C}_w^{\mathbf{R}}{\mathbf{X}}^i_{w-1}
    \\
    &\kern -0.2cm\mathbf{Y}^i_w = \chi\left(\mathsf{softmax}\left(\frac{\beta(\mathbf{Q}^i_w)\beta((\mathbf{K}^i_w)^{\top})}{\sqrt{|\mathcal{N}^{i,k}|}}\right) \beta(\mathbf{V}^i_w)\right),
    \\
    &\kern -0.2cm\mathbf{X}^i_w = \psi(\mathbf{D}_w^{\mathbf{R}}{\mathbf{Y}}_w^i),
\end{aligned}
\end{gather}
where $\beta(\cdot)$, $\chi(\cdot)$, and $\psi(\cdot)$ are nonlinear activation functions. In the aforementioned operations, $\mathbf{A}_w^{\mathbf{R}}, \mathbf{B}_w^{\mathbf{R}}, \mathbf{C}_w^{\mathbf{R}} \in \mathbb{R}^{r_w \times h_w}$ and $\mathbf{D}_w^{\mathbf{R}} \in \mathbb{R}^{n_x \times r_w}$ for $w =1, \ldots, W$ are matrices to be learned and shared across robots; and $h_w, r_w, d_w>0$, with $d_W = n_x$ and $h_1 = n_x$ for valid matrix multiplications. Matrices $\mathbf{A}_w^{\mathbf{R}}, \mathbf{B}_w^{\mathbf{R}}, \mathbf{C}_w^{\mathbf{R}}$ are of fixed size, so they are independent of the number of robots and neighbors. Thus, robot $i$ can deal with time-varying neighbors. In particular, $\mathbf{A}_1^{\mathbf{R}}, \mathbf{B}_1^{\mathbf{R}}, \mathbf{C}_1^{\mathbf{R}}$ transforms the states to features encoded in the query $\bfQ_1^i$, key $\bfK_1^i$ and value $\bfV_1^i$ matrices. 
Then, $[\mathbf{R}_{\bftheta}(\bfx)]_{ij}$ is constructed as a weighted matrix that models the interactions of robot $i$ with its $k$-hop neighbors, and a diagonal positive semidefinite matrix that accounts for the self-interactions:
\begin{gather}\label{eq:Restimate}
\begin{aligned}
\mathbf{Z}_{ij}^{\bfR} &= \mathsf{diag}(\mathbf{x}^{i,j}_W),\\
[\mathbf{R}_{\bftheta}(\bfx)]_{ij} &= -({\mathbf{Z}}_{ij}^{\bfR}+{\mathbf{Z}}_{ji}^{\bfR}), \quad  \forall j \in \mathcal{N}^{i,k},\\
[\mathbf{R}_{\bftheta}(\bfx)]_{ii} &= \mathbf{Z}_{ii}^{\bfR} +  \sum_{j \in \mathcal{N}^{i,k}}   (\mathbf{Z}_{ij}^{\bfR}+{\mathbf{Z}}_{ji}^{\bfR}),    
\end{aligned}
\end{gather}
where $\mathbf{x}^{i,j}_W$ is the column that corresponds to neighbor $j$ in $\mathbf{X}^i_W$, and  $\mathsf{diag}(\cdot)$ is the operator that reshapes the $n_x \times 1$ vector to a $n_x \times n_x$ diagonal matrix. If the nonlinear activation function $\psi(\cdot)$ is designed such that the elements of the output $\bfX_W^i$ are positive, then $\mathbf{Z}_{ij}^{\bfR}$ is a diagonal positive semidefinite matrix and $\mathbf{R}_{\bftheta}(\bfx)$ is a diagonally dominant matrix. This way, $\mathbf{R}_{\bftheta}(\bfx)$ is positive semidefinite by design. In fact, note the similarities between \eqref{eq:Restimate} and a weighted Laplacian matrix. More importantly, each element $[\mathbf{R}_{\bftheta}(\bfx)]_{ij}$ depends only on the information from robot $i$ and its neighbor $j$, so the computation is distributed.

To learn $[\mathbf{J}_{\bftheta}(\bfx)]_{ij}$, we follow the same steps in \eqref{eq:R_layers}, with parameters $\mathbf{A}_w^{\mathbf{J}}, \mathbf{B}_w^{\mathbf{J}}, \mathbf{C}_w^{\mathbf{J}},\mathbf{D}_w^{\mathbf{J}}$ instead of $\mathbf{A}_w^{\mathbf{R}}, \mathbf{B}_w^{\mathbf{R}}, \mathbf{C}_w^{\mathbf{R}},\mathbf{D}_w^{\mathbf{R}}$, to obtain encodings $\bfZ_{ij}^\bfJ$ instead of $\bfZ_{ij}^\bfR$. Due to the reciprocal communication between robots $i$ and $j$, we enforce the skew-symmetry of $\bfJ_\bftheta(\bfx)$ by:
\begin{equation}\label{eq:Jestimate}
    [\mathbf{J}_{\bftheta}(\bfx)]_{ij} = {\mathbf{Z}}_{ij}^{\bfJ}-{\mathbf{Z}}_{ji}^{\bfJ} \quad \forall j \in \mathcal{N}^{i,k}.
\end{equation}
Since $[\mathbf{J}_{\bftheta}(\bfx)]_{ii} = \mathbf{0}$, the interconnection matrix is such that $\bfJ_\bftheta(\bfx) + \bfJ_\bftheta^\top(\bfx)=\bf0$ and, thus, is skew-symmetric by design. Again, each element $[\mathbf{J}_{\bftheta}(\bfx)]_{ij}$ depends only on the information from robot $i$ and its neighbor $j$, so the computation is distributed.

Finally, to learn $H^{i}_{\bftheta}(\bfx^i_{\calN^{i,k}})$, we represent it as follows:
\begin{equation}\label{eq:Hestimate}
H^{i}_{\bftheta}(\bfx^i_{\calN^{i,k}}) = \mathsf{vec}(\bfx^i_{\calN^{i,k}})^{\top} \mathbf{M}_{\bftheta}^{i}(\bfx^i_{\calN^{i,k}})\mathsf{vec}(\bfx^i_{\calN^{i,k}}) + {U}_{\bftheta}^{i}(\bfx^i_{\calN^{i,k}}).
\end{equation}
The first term in the right-hand side of Eq.~\eqref{eq:Hestimate} is a kinetic-like energy function and the second terms is a potential energy function with $$\mathbf{M}_{\bftheta}^{i}(\bfx^i_{\calN^{i,k}}) = \diag( \mathbf{1}^{\top}\mathbf{Z}^\bfM_i) \text{ and }{U}_{\bftheta}^{i}(\bfx^i_{\calN^{i,k}}) = \mathbf{1}^{\top}\mathbf{Z}^U_i\mathbf{1}.$$
The encodings $\mathbf{Z}^\bfM_i$ and $\mathbf{Z}^U_i$ are calculated using the same steps in \eqref{eq:R_layers}, with parameters $\mathbf{A}_w^{\mathbf{M}}$, $\mathbf{B}_w^{\mathbf{M}}$, $\mathbf{C}_w^{\mathbf{M}}$, $\mathbf{D}_w^{\mathbf{M}}$ and $\mathbf{A}_w^{{U}}$, $\mathbf{B}_w^{{U}}$, $\mathbf{C}_w^{{U}}$, $\mathbf{D}_w^{{U}}$, respectively.
Once we have $H^{i}_{\bftheta}(\bfx^i_{\calN^{i,k}})$, we obtain $\partial H^{j}_{\bftheta}(\bfx^j_{\calN^{j,k}})/\partial \bfx^i$ from all the neighboring robots and compute  Eq.~\eqref{eq:partial_partial}. Therefore, since all the operations are only dependent on the available information in robot $i$ and its neighbor $j$, the computations involving the Hamiltonian function are distributed by design.

To deploy the control policy \eqref{eq:open_loop_intro_controller_ind_distributed}, we design a message $\mathbf{m}^{i,j}_\tau$, encoding information that robot $i$ needs from robot $j$ at time $\tau$ to calculate $[\mathbf{J}_{\bftheta}(\bfx)]_{ij}, [\mathbf{R}_{\bftheta}(\bfx)]_{ij}$, and $H_{\bftheta}^{i}(\bfx)$. When there is no communication among robots, i.e., $k=0$, no message is needed. For $k \geq 1$, robot $i$ uses the following communication protocol:
\begin{enumerate}
    \item Robot $i$ receives messages $\mathbf{m}_{\tau,1}^{i,j} = \bfx_j$ from its $k$-hop neighbors in $\calN^{i,k}$. Then, robot $i$ computes ${\mathbf{Z}}_{ij}^{\bfR}$, ${\mathbf{Z}}_{ij}^{\bfJ}$, $H_{\bftheta}^{i}$, and $\partial H^{i}_{\bftheta}/\partial \bfx_j$.
    
    \item Robot $i$ receives messages $\mathbf{m}_{\tau,2}^{i,j} = \{\partial H^{j}_{\bftheta}/\partial \bfx_i, {\mathbf{Z}}_{ji}^{\bfR}, {\mathbf{Z}}_{ij}^{\bfJ}\}$ from its $k$-hop neighbors in $\calN^{i,k}$ and computes $\partial H_{\bftheta}/\partial \bfx_i$, $[\mathbf{J}_{\bftheta}]_{ij}, [\mathbf{R}_{\bftheta}]_{ij}$.
    
    \item Robot $i$ receives messages $\mathbf{m}_{\tau,3}^{i,j} = \partial H_{\bftheta}/\partial \bfx_j$ from its $k$-hop neighbors in $\calN^{i,k}$ and computes the control input $\bfa_i$.
\end{enumerate}
In summary, each robot $i$ receives a message \mbox{$\mathbf{m}_{\tau}^{i,j} = [\mathbf{m}_{\tau,1}^{i,j},\mathbf{m}_{\tau,2}^{i,j}, \mathbf{m}_{\tau,3}^{i,j}]$} in $3$ communication rounds from its neighboring robot $j$. We assume negligible delays between communication rounds. If the delay is large, Wang et al.~\cite{wang2022darl1n} suggest to learn a function that predicts quantities such as $\partial H_{\bftheta}(\bfx) / \partial \bfx_j$, ${\mathbf{Z}}_{ji}^{\bfJ}$, ${\mathbf{Z}}_{ji}^{\bfR}$, leading to one communication round. We leave this for future work. 

\section{Physics-informed multi-robot soft actor-critic}\label{sec:sac}

\begin{figure*}[t]
    \centering
    \includegraphics[width=\linewidth, trim={0 8.5cm 0 0}, clip]{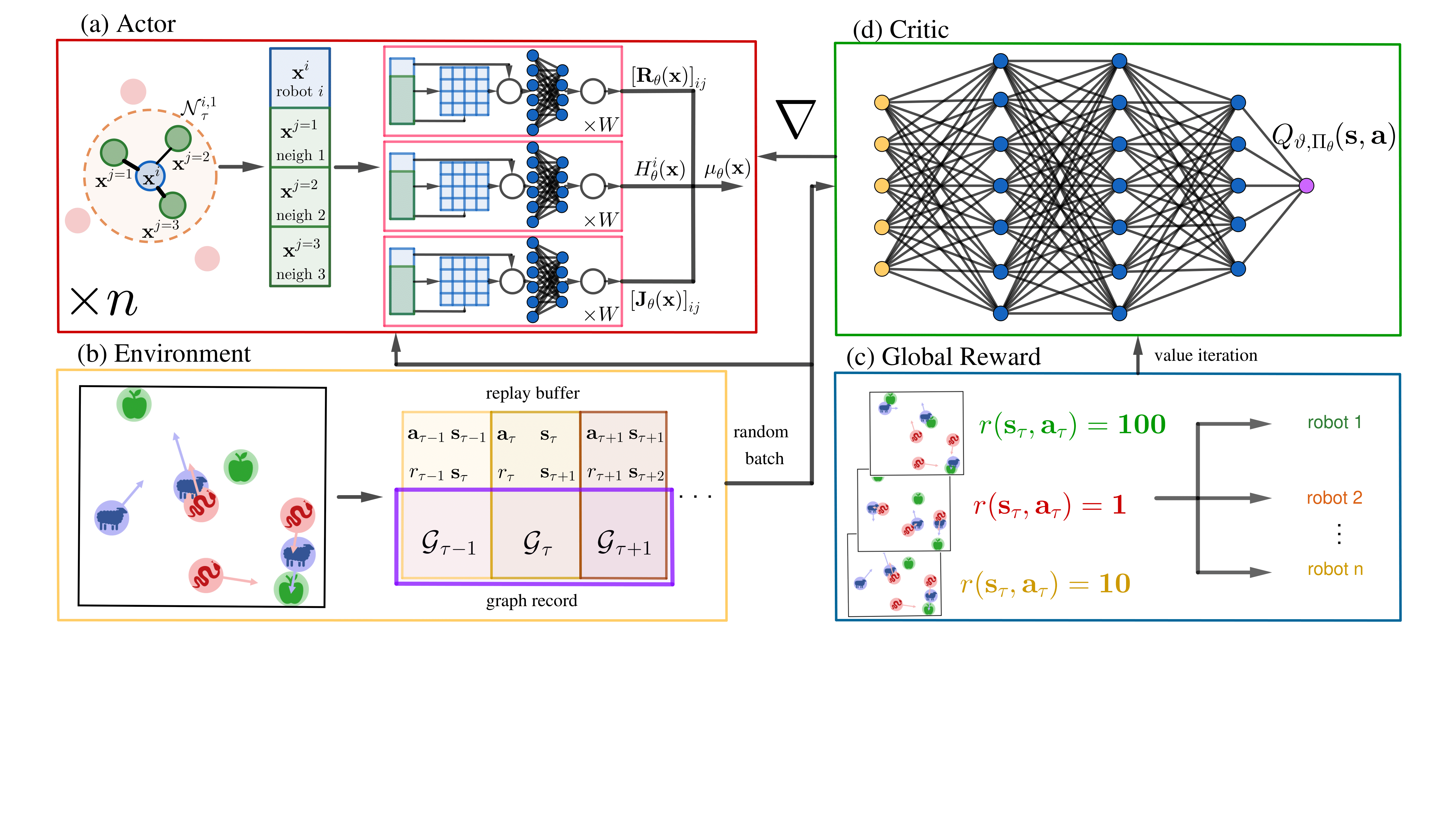}
    \caption{\small{{Overview of our physics-informed soft actor-critic multi-agent reinforcement learning approach. The main differences with respect of other actor-critic methods are the following: (a) the actor is the multi-robot network modeled as a port-Hamiltonian system with a self-attention-based IDA-PBC policy; (b) the replay buffer not only stores actions, states and rewards, but also the graph structure of the multi-robot network to enforce the desired distributed structure; (c) the reward is global because the actor is the whole multi-robot team and the physics-informed parameterization conditions the policy on the graph structure of the team; (d) the output of the critic is shared across robots and the policy parameters are the same for all robots, so for the same critic gradient step, $n$ policy gradient steps are taken.}}}
    \label{fig:overall_pipeline}
\end{figure*}

The previous section describes how to parameterize the distributed control policies to be learned. In this section, we present a soft actor-critic algorithm \cite{haarnoja2018soft2} to train the self-attention port-Hamiltonian neural network, detailing the main features that allow the integration of our physics-informed policy representation. 
The key components are presented in Fig.~\ref{fig:overall_pipeline}, namely: (a) actor, (b) environment, (c) reward, and (d) critic.

\subsection{The actor}\label{subsec:actor}

First, we model the actor with the port-Hamiltonian system detailed in Sec.~\ref{subsec:portHamiltonian} and we use the IDA-PBC policy in Eq.~\eqref{eq:open_loop_intro_controller_ind_distributed} parameterized by the self-attention architecture proposed in Sec.~\ref{subsec:LEMURS}. The known robot dynamics are typically provided by the simulation environments or the hardware specifications of the robots~\cite{bettini2022vmas}. 
Nevertheless, two aspects must be adapted to match the soft actor-critic formulation: (i) the MDP formulation and the stochastic Markov policy are in discrete time while the robot dynamics and the IDA-PBC are in continuous-time, and (ii) the actions in the soft actor-critic method are stochastic while the port-Hamiltonian formulation is deterministic.

To address the continuous versus discrete time mismatch, we use the zero-order hold control approach described in Sec.~\ref{sec:prosta}. More specifically,
$\bfu(t) = \bfa_\tau = \bfmu_{\bftheta}(\bfx(t_\tau))$ for $t \in [t_\tau, t_{\tau+1})$.
This means that we can write the control policy as dependent on the discrete-time state $\bfs_\tau$ rather than the continuous-time state $\bfx(t_\tau)$, leading to $\bfmu_{\bftheta}(\bfs_\tau)$ with the same expression in Eq. \eqref{eq:open_loop_intro_controller_ind_distributed}.

To address stochasticity and maintain the desired distributed structure already derived in Sec.~\ref{subsec:portHamiltonian} and \ref{subsec:LEMURS}, we model the distributed control policies as (squashed) Gaussian distributions, whose mean is given by the IDA-PBC controller in \eqref{eq:open_loop_intro_controller_ind_distributed}, therefore respecting the distributed policy factorization. Meanwhile, the variance of the policy distribution is provided by a neural network that is learned during training. Overall, we obtain a distributed and stochastic Markov control policy of the form:
\begin{equation}\label{eq:squashed_normal}
  \mathbf{a}_\tau^i \sim \bfpi_\bftheta(\mathbf{a}_\tau^i|\mathbf{s}^i_{\calN_\tau^{i,k}}) = \tanh(\bfmu_\bftheta(\mathbf{s}^i_{\calN_\tau^{i,k}}) + \bfsigma_\varrho(\mathbf{a}_\tau^i, \mathbf{s}^i_{\calN_\tau^{i,k}})  \xi), 
\end{equation}
where $\bfmu_\bftheta(\mathbf{s}^i_{\calN_\tau^{i,k}})$ is the IDA-PBC policy parameterized by self-attention neural networks under the restrictions imposed by the networked structure of the robot team.

On the other hand, $\bfsigma_\varrho(\mathbf{a}_\tau^i, \mathbf{s}^i_{\calN_\tau^{i,k}})$ is a vector of standard deviations given by a neural network that approximates the variance of the squashed Gaussian distribution with parameters $\varrho$. Besides, $\xi \sim \calN(\mathbf{0}, \mathbf{I})$, so the control policy is a Gaussian distribution with mean $\bfmu_\bftheta(\mathbf{s}^i_{\calN_\tau^{i,k}})$ and diagonal covariance matrix whose diagonal is equal to $(\bfsigma_\varrho(\mathbf{a}_\tau^i, \mathbf{s}^i_{\calN_\tau^{i,k}}))^2$. The control is constrained to $\mathbf{a}_\tau^i \in [-1, 1]^{n_u}$ by means of a $\tanh$ function \cite{haarnoja2018learning, haarnoja2018soft, haarnoja2018soft2}, leading to a squashed Gaussian policy. In practice, the control input can be constrained to $\mathbf{a}_\tau^i \in [a_{\min}, a_{\max}]^{n_u}$ with $-\infty < a_{\min} \leq a_{\max} < \infty$ by scaling the output of the $\tanh$ function.
It is also important to remark that $\bfsigma_\varrho(\mathbf{a}_\tau^i, \mathbf{s}^i_{\calN_\tau^{i,k}})$ only depends on the available information at each robot, such that the desired distributed factorization of the control policy is preserved.

During training, the control inputs are sampled from the Gaussian distribution, as stated in Eq.~\eqref{eq:general_controller}. After training, the mean control input is chosen, i.e., $\mathbf{a}^i_\tau = \tanh(\bfmu_\bftheta(\mathbf{s}^i_{\calN_\tau^{i,k}})).$ In the case that an additional layer of robust, adaptive or active control is desired, then $\bfsigma_\varrho(\mathbf{a}_\tau^i, \mathbf{s}^i_{\calN_\tau^{i,k}})$ can be used as a proxy of the uncertainty in the control policy, since it only depends on the neighboring information at robot $i$. The design of $\bfsigma_\varrho(\mathbf{a}_\tau^i, \mathbf{s}^i_{\calN_\tau^{i,k}})$ is free to choose, but in this work we opt for the same architecture in Eq.~\eqref{eq:R_layers}. 



\subsection{The environment}\label{subsec:environment}

The second modification over the soft actor-critic algorithm is in the collection of experiences from the environment. The environment is determined by the desired task to be solved, with examples found in Fig.~\ref{fig:overall_scenarios}. The main difference with other multi-agent reinforcement learning works is that, to build the replay buffer that stores the trial and error experience used for training the control policies, we take into account all the interactions among the robots. Differently from centralized-training decentralized-deployment approaches, where the experience of each agent is recorded independently of the other agents, we record for each experience all the robot states, actions, and interaction graph together.
Thanks to that, we keep track of the correlation among robots during training. This also allows us to condition the trained control policies on the available information at each robot, therefore only providing the information robots will have access during deployment.

\subsection{The reward}\label{subsec:reward}

The reward function can be shared across robots and include global terms because the actor is the whole multi-robot network. This is important because, from the perspective of the soft actor-critic algorithm during training, the whole robot network is a single centralized agent. Other state-of-the-art multi-agent reinforcement learning approaches (see Sec.~\ref{sec:related}) require a factorization of the reward function because each agent is considered as an isolated learning unit which does not exploit neighboring information during the execution of the control policy. In contrast, in our approach the shared distributed control policy is simultaneously learned at all the robots used in training, and since the experience records the underlying communication graph, we can seamlessly associate global rewards with distributed cooperative control policies through communication. It is worth to note that, as in any classical soft actor-critic approach, the maximization objective is changed to include an entropy term
$\calH(\bfPi_{\bftheta}(\bfa|\mathbf{s}))$ that measures the entropy of the control policy. This entropy term, weighted by a temperature parameter $\alpha>0$, trades off exploration (high value) and exploitation (low value). The value of $\alpha$ changes over time according to a gradient descend law which manages automatically the exploration/exploitation dichotomy.
The details of this automatic temperature adjustment rule can be found in \cite{haarnoja2018soft2}.

\subsection{The critic}\label{subsec:critic}

Finally, regarding the critic, its main purpose is to learn the action-value function of the environment, encoded in $Q_{\bfPi_\bftheta}(\bfs, \bfa)$. This approximation steers the training of the control policy towards the maximization objective. Nevertheless and importantly, the critic is only used during training. Thus, it is not necessary to design it to be distributed and an existing centralized neural network architecture can be used. In this work, we opt for a multi-layer perceptron with parameters $\bm{\vartheta}$ to learn $Q_{\bm{\vartheta}, \bfPi_\bftheta}(\bfs, \bfa)$. As with the reward, since, from the perspective of the soft actor-critic algorithm, the single agent is the whole robot network described by the port-Hamiltonian dynamics, a single centralized critic can learn the appropriate action-value function to condition the training of the distributed control policy.

A shared centralized action value function implies that each optimization step considers, simultaneously, all the policies gathered in the joint policy $\bfPi_\bftheta$. Since the policies are homogeneous, each optimization step is simultaneously updating the policy $\bfpi_\bftheta$ $n$ times. This is achieved without any particular specification of the gradients nor factorization of the action value function. This is one of the key properties and advantages of our proposed soft actor-critic algorithm compared to other state-of-the-art algorithms. Typical multi-agent reinforcement learning approaches, by factorizing the policy and/or the value functions, arrive to a different set of policy parameters, one per agent used in training. In cooperative tasks where the robots share the same goal, there should be a single policy that resolves the task independently on the configuration of the robot. In our case, by appropriately integrating the modular port-Hamiltonian description of the multi-robot system with the soft actor-critic algorithm, we train a single distributed control policy which takes into account the current available information at each robot. Using the soft actor-critic update rules \cite{haarnoja2018soft, haarnoja2018soft2, haarnoja2018learning}, the gradients with respect to $\bftheta$ are, simultaneously for all the robots, already conditioned on the multi-robot communication topology.

In conclusion, in contrast with other solutions, ours explicitly considers the exchanges of information among robots by modeling the multi-robot system as a graph. This allows to use a reinforcement learning algorithm for single-robot problems, where the single agent is the multi-robot network. From the perspective of the reinforcement learning algorithm, the actor is centralized. However, by means of a physics-informed self-attention parameterization of the dynamics and control of the robots, the learned policies are distributed by design, modeling the robot team as a modular port-Hamiltonian system.


\section{Results}\label{sec:simulations}

To assess our physics-informed multi-agent reinforcement learning approach, we present {seven} multi-robot scenarios. The first three are extracted from the VMAS simulator \cite{bettini2022vmas}. The next three are adaptations from the MPE simulator \cite{mordatch2018emergence, lowe2017multi} that can be found in \cite{wang2022darl1n}. {The last scenario comes from the Multi-Agent MuJoCo benchmark \cite{peng2021facmac}. The overview of the scenarios are detailed in the following}:\renewcommand{\theenumi}{\alph{enumi}}
\begin{enumerate}
    \item \textbf{Reverse transport}: the robots are randomly spawned inside a box that they must push towards a desired landmark in the arena. The initial position of the box and the landmark is random. Compared to \cite{bettini2022vmas}, we decrease the mass of the box to $1$ kg to ensure that the box can be moved even with a small number of robots. Also, when the box is in the landmark, the reward is set to $1$.
    
    \item \textbf{Sampling}: the robot team is randomly spawned in an arena with an underlying Gaussian density function composed of $3$ modes. The field is discretized to a grid. Robots must collect samples of the field such that once a robot visits a cell its sample is collected without replacement and given as reward to the team. Robots use a LiDAR to sense each other, and they observe the samples in the $3\times3$ grid around it. 
    
    \item \textbf{Navigation}: each robot has a landmark to reach. The initial position of the robots and landmarks are randomly spawned in a $2\times2$m square arena. Robots must navigate to reach their corresponding landmarks while avoiding collisions with other robots. Compared to \cite{bettini2022vmas}, we encourage collision avoidance by changing the collision penalty from $-1$ to $-5$. Besides, each robot only observes its desired landmark instead of all the landmarks. 
    
    \item \textbf{Food collection}: it is a version of the simple spread scenario \cite{lowe2017multi} where a team of robots and food landmarks is randomly spawned in an arena. There are as many landmarks as robots. Robots must cooperate to cover as many food landmarks as possible. Each time a robot covers a new landmark, the whole team is rewarded. 
    
    \item \textbf{Grassland}: A team of robots must collect food resources that are randomly spawned in an arena, while evading multiple predators. There are as many robots as predators, and the former move twice faster than the latter. The predator team is positively rewarded when some member captures a robot, whereas the robot team is negatively rewarded and the robot is deactivated. On the other hand, the robots receive a positive reward if they reach a food landmark, which is then spawned again in a new position.
    
    \item \textbf{Adversarial}: two robotic teams compete for the same food landmarks. Both teams have the same number of robots. When a member of a team reaches a food landmark, the landmark is randomly spawned and the team is positively rewarded. When two members of a team collide with one member of the other team, then the first team is positively rewarded, while the second team is penalized and the robot deactivated.

    \item {\textbf{6x1-Half Cheetah}: a robotic cheetah with two legs and 6 joints, where each joint is a different agent. The multi-agent team is distributed in the sense that each agent can communicate only with the adjacent joints, enforced by a ring graph topology. The task is to make the robot run as fast as possible. The reward is global and is composed by a first term that favours forward movements and a second term that penalizes too large actions.}
\end{enumerate}
The first three scenarios are used for ablation studies, {where the main goal is to evaluate how our physics-informed policy parameterization improves upon existing standard policy parameterizations. The second three are used to compare our method with other state-of-the-art multi-agent reinforcement learning approaches that are not physics-informed. The last scenario is used to validate our approach in a realistic robotic platform.}

All robots have a communication radius $r_{comm} > 0$ which allows to exchange information with $1$-hop neighbors. The communication radius and other hyperparameters of the soft actor-critic algorithm are specified in Appendix~\ref{sec:sac_parameters}. Each robot observes its position, velocity, position and velocity of the landmarks or objects of interest (e.g., the box in reverse transport). The particularities of the observation space of each robot can be found in \cite{bettini2022vmas, lowe2017multi, wang2022darl1n}. In some scenarios, the robot observation vector changes its dimension depending on the number of robots. Specifically, in the food collection, grassland and adversarial scenarios the number of landmarks changes with the number of robots. Therefore, to accommodate the proposed self-attention port-Hamiltonian neural network with an observation vector which may change its size, we use an additional neural network to pre-process the observation vector. In particular, we concatenate a self-attention layer with a dense layer which receives, as input, the food landmarks' positions and outputs a feature vector of constant dimension that is used as part of the state vector to build $\mathbf{S}_t^i$. The details can be found in Appendix~\ref{sec:parameters}.
Supplementary material can be found in our repository\footnote{ \url{https://github.com/EduardoSebastianRodriguez/phMARL}} {and the supplementary video\footnote{{\url{https://youtu.be/pSzP3LBVyZg}}}}.

\begin{figure*}
    \centering
    \begin{tabular}{ccc}
(a) Reverse transport
    & 
(b) Sampling  
&
(c) Navigation  
    \\
\includegraphics[width=0.31\textwidth, height=0.2\textwidth]{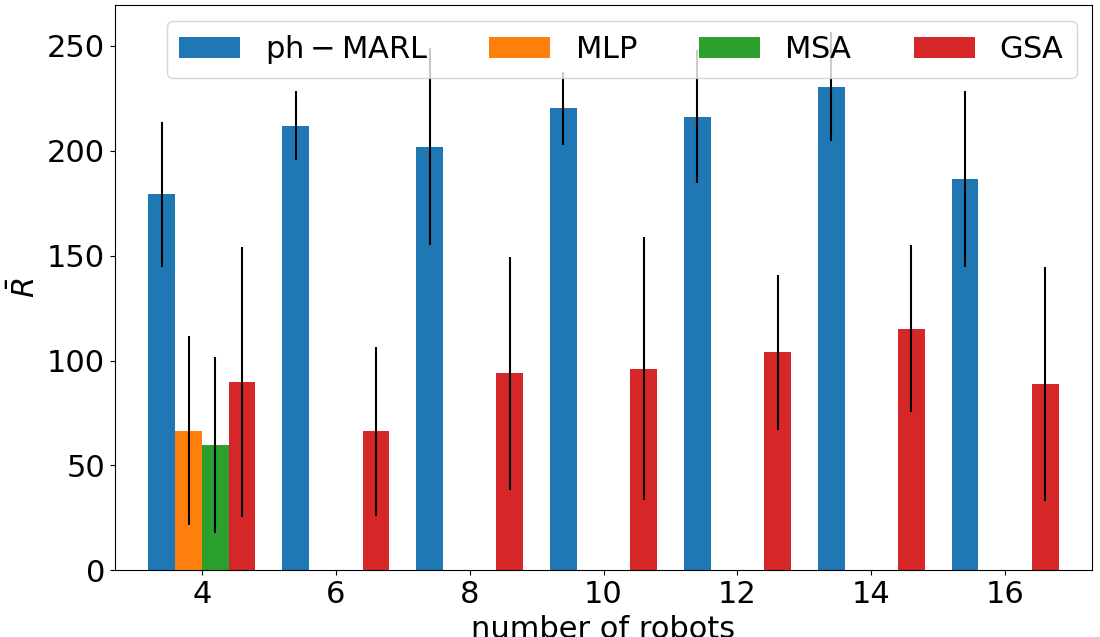}
    & 
\includegraphics[width=0.31\textwidth, height=0.205\textwidth]{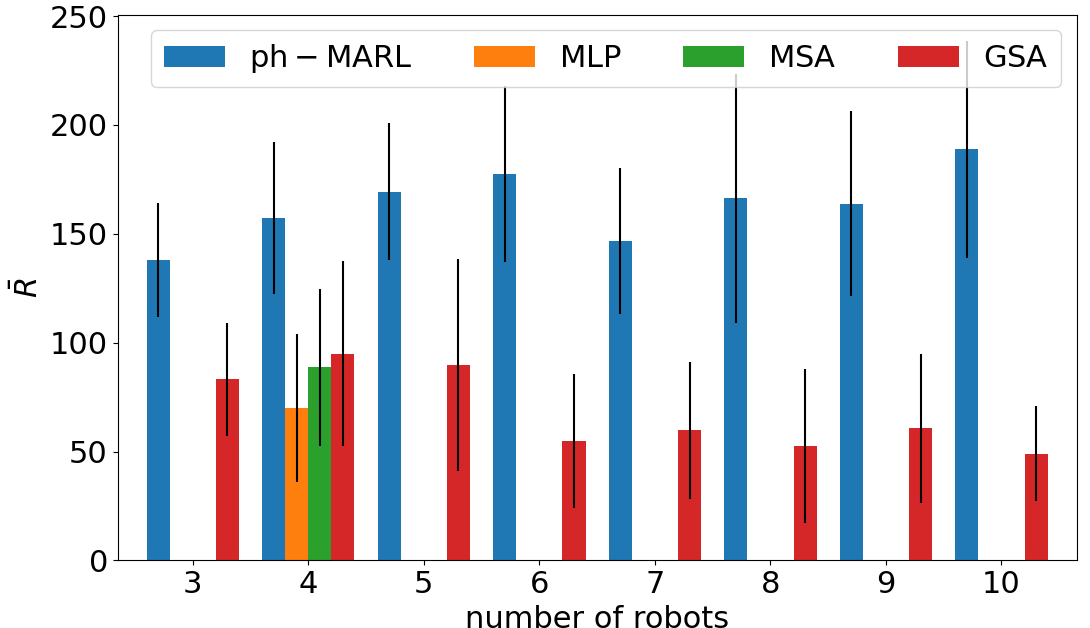}  
&
\includegraphics[width=0.31\textwidth, height=0.2\textwidth]{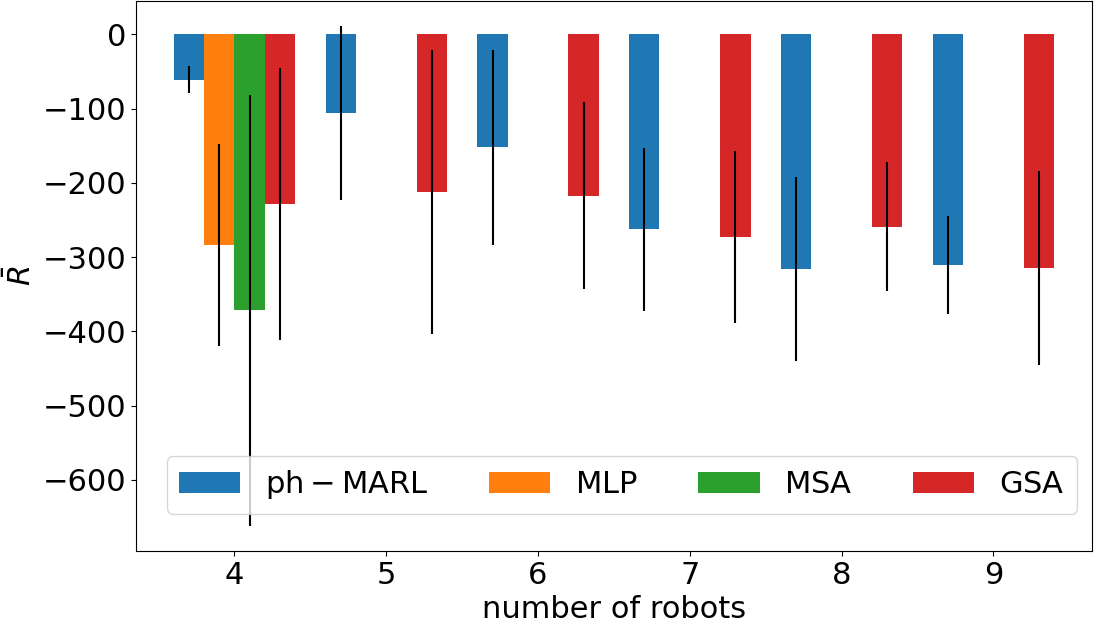}
    \end{tabular}
    \caption{Comparison of the performance of the ablated control policies when we scale the number of robots in deployment. In all the scenarios, our proposed combination of a port-Hamiltonian modeling and self-attention-based neural networks achieves the best cumulative reward without further training the control policy. Each bar displays the mean and standard deviation of $\Bar{R}$ over $10$ evaluation episodes.}
    \label{fig:scalability_results_ablation}
\end{figure*}

\begin{figure*}
    \centering
    \setlength\tabcolsep{14pt}
    \begin{tabular}{ccccc}
&
$n = 4$ robots
&
$n = 8$ robots
&
$n = 12$ robots
&
$n = 16$ robots
\\
\raisebox{40pt}{\rotatebox[origin=c]{90}{(a) Reverse transport}}
    &
\includegraphics[width=0.18\textwidth, height=0.18\textwidth]{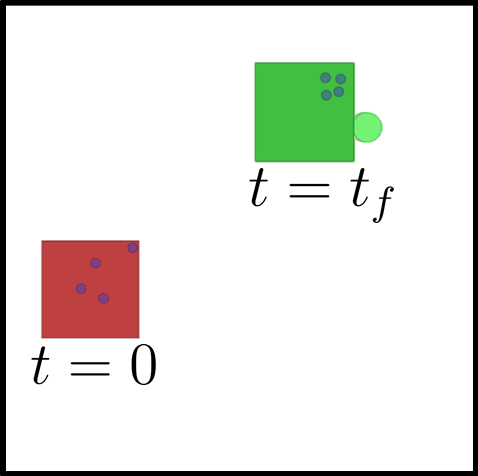}
    & 
\includegraphics[width=0.18\textwidth, height=0.18\textwidth]{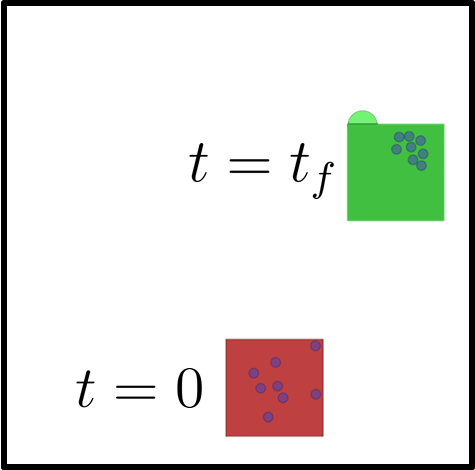}
    & 
\includegraphics[width=0.18\textwidth, height=0.18\textwidth]
{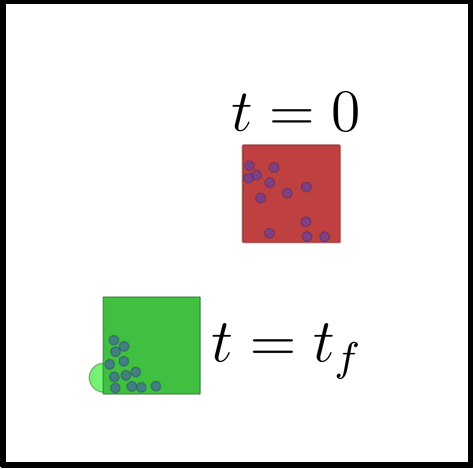} 
    & 
\includegraphics[width=0.18\textwidth, height=0.18\textwidth]{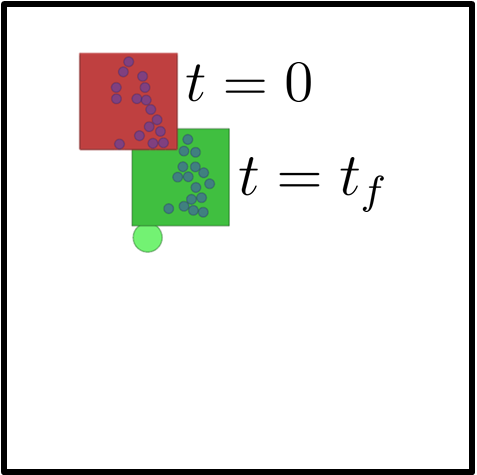}
\\
&
$n = 3$ robots
&
$n = 5$ robots
&
$n = 7$ robots
&
$n = 9$ robots
\\
\raisebox{40pt}{\rotatebox[origin=c]{90}{(b) Sampling}}
    &
\includegraphics[width=0.18\textwidth, height=0.183\textwidth]{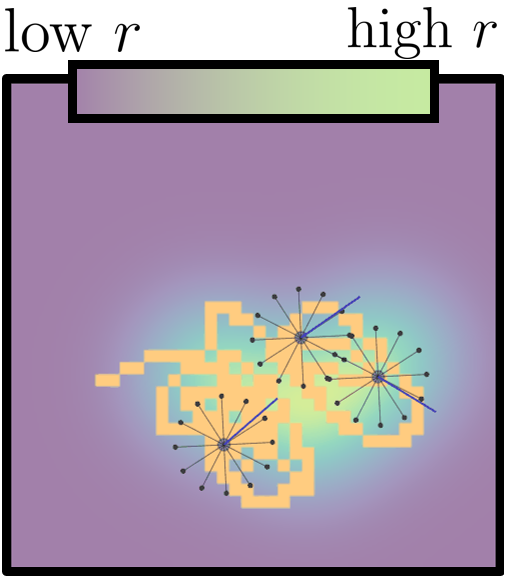}
    & 
\includegraphics[width=0.18\textwidth, height=0.183\textwidth]{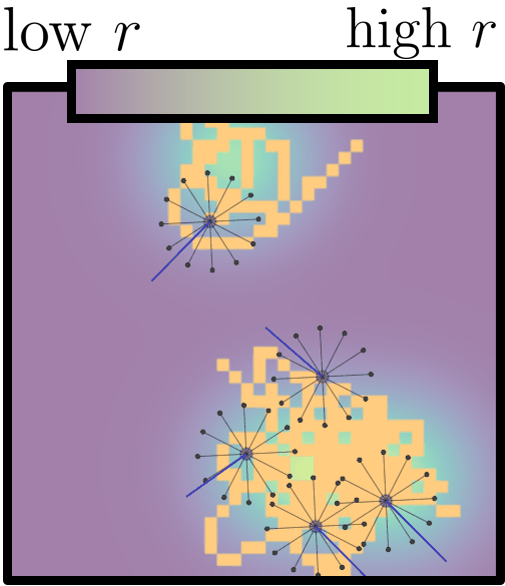}
    & 
\includegraphics[width=0.18\textwidth, height=0.1825\textwidth]
{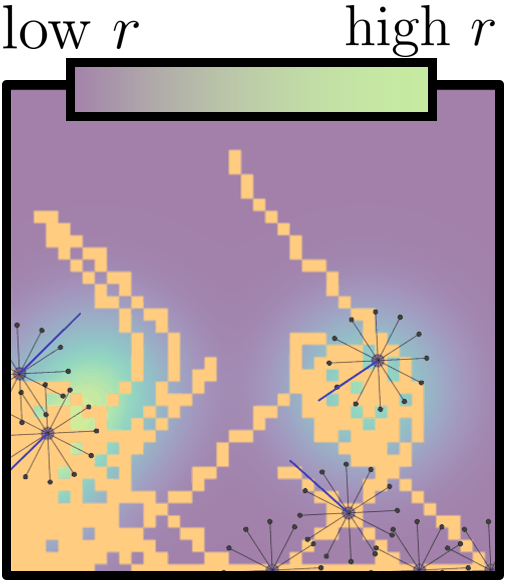} 
    & 
\includegraphics[width=0.18\textwidth, height=0.1825\textwidth]{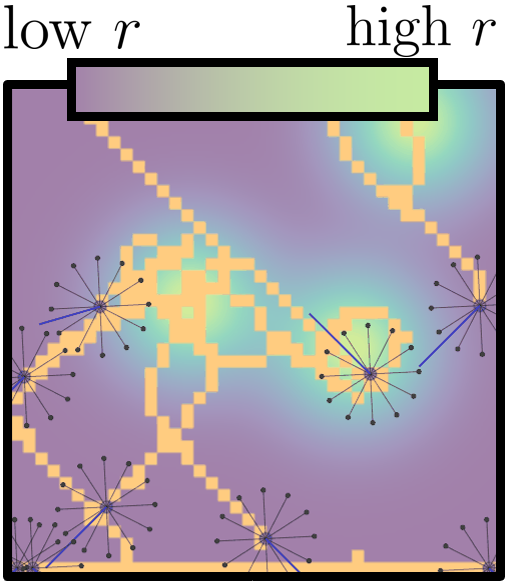}
\\
&
$n = 4$ robots
&
$n = 5$ robots
&
$n = 6$ robots
&
$n = 8$ robots
\\
\raisebox{40pt}{\rotatebox[origin=c]{90}{(c) Navigation}}
    &
\includegraphics[width=0.18\textwidth, height=0.183\textwidth]{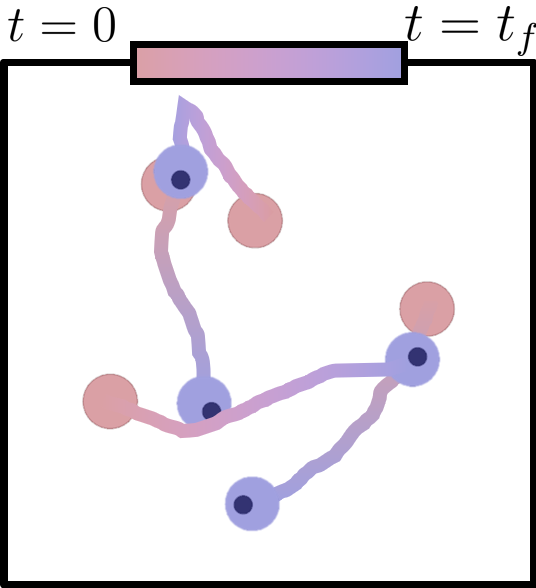}
    & 
\includegraphics[width=0.18\textwidth, height=0.183\textwidth]{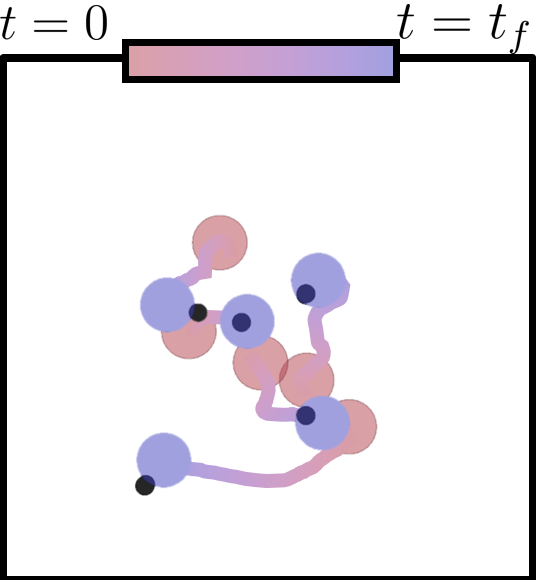}
    & 
\includegraphics[width=0.18\textwidth, height=0.183\textwidth]
{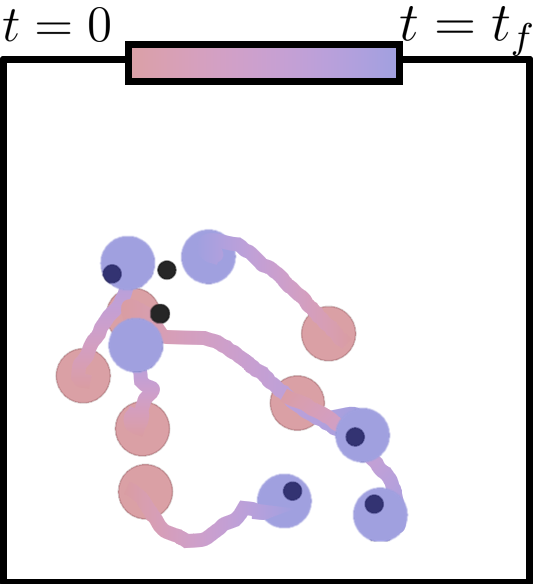} 
    & 
\includegraphics[width=0.18\textwidth, height=0.183\textwidth]{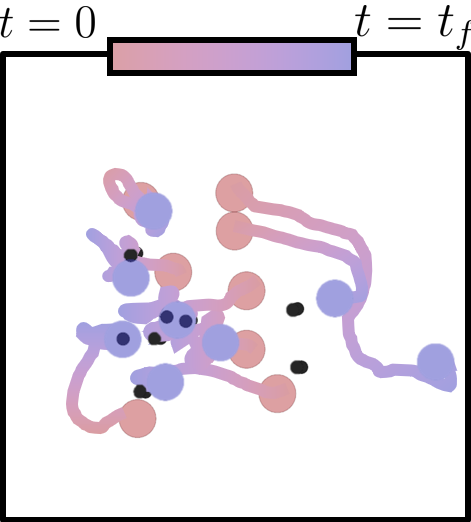}  
    \end{tabular}
    \caption{Examples of multi-robot scenarios for different initial conditions and numbers of robots. {In the reverse transport scenario, the red box turns green when it reaches the desired goal (green dot), and the robots are depicted as small blue dots.} {In the sampling scenario, LiDAR measurements are depicted as points forming a star pattern centered at the position of the robot, and regions already explored are shown in orange.} {In the navigation scenario, evolution over time of the trajectories is colored from pink ($t=0$) to blue ($t=t_f$), whereas the goals are depicted as black dots.} It is interesting to see that some aspects of the environment do not scale with the team size, e.g., the size and weight of the box in the reverse transport scenario, the number of hot spots in the sampling scenario or the size of the arena in the navigation scenario.}
    \label{fig:instances_results_ablation}
\end{figure*}

\subsection{Ablation results}\label{subsec:ablation}

We conduct ablation studies using the reverse transport, sampling and navigation scenarios. {The goal is to evaluate whether our physics-informed policy parameterization improves upon existing standard policy parameterizations.} We compare the proposed self-attention port-Hamiltonian neural network (\textsf{pH-MARL}) with a classical multi-layer perceptron (\textsf{MLP}), a modular self-attention-based neural network (\textsf{MSA}) and a graph- and attention-based neural network (\textsf{GSA}). The implementation details can be found in Appendix \ref{sec:parameters}. 

The \textsf{MLP} network is an unstructured neural network that receives, as input, the current states and actions of the robots and outputs the action vector. The \textsf{MSA} network replaces the dense layers of the \textsf{MLP} network with an architecture based on self-attention, but has the same input and output. The \textsf{GSA} network is similar to the \textsf{MSA} network but has the adjacency matrix as additional input. The three neural networks constitute a sequential improvement from a standard neural network to our port-Hamiltonian formulation. It is important to note that, with the current formulation, it is not possible to try a port-Hamiltonian neural network which is not based on self-attention because each element of the processed input sequence is employed to compute the port-Hamiltonian terms, that are also a sequence spanned from $i \in \calN_t^{i,k}$. This is key for the scalability of the policy and is a feature that is not provided by other architectures based on, e.g., multi-layer perceptrons or convolutions. 

Finally, we employ the same architecture used in \textsf{pH-MARL} to learn the standard deviation for all the options, along with the same critic and soft actor-critic algorithm  (Appendix~\ref{sec:parameters}).


Table~\ref{tab:converge_R} shows the averaged cumulative reward $\boldsymbol{\Bar}{R} = \frac{1}{n} R = \frac{1}{n}\sum_{\tau=0}^{\tau_{\max} - 1}r(\bfs_\tau,\bfa_\tau)$ after convergence of the training, where $\tau_{\max}$ is the maximum number of steps per episode. For the three scenarios and the four neural network architectures, we use $n=4$ robots. The table reports the mean of $\Bar{R}$ over $10$ evaluation episodes.  In all the cases, \textsf{pH-MARL} surpasses the other three architectures, with a particularly significant difference in the reverse transport and navigation scenarios. Given the same number of parameters in the four architectures, the use of a physics-informed formulation of the neural network leads to a structured learning with efficient sampling in training. The difference in the reverse transport scenario is due to the fact that the other networks sometimes fail in reaching the landmark, despite arriving to a very close position. In the sampling scenario, \textsf{pH-MARL} is the fastest in inspecting the environment, therefore covering more informative cells. It is followed, in decreasing order, by \textsf{GSA}, \textsf{MSA} and \textsf{MLP}, which is reasonable since the three neural networks are in decreasing order of architecture complexity in terms of neural network modules and input information. The navigation scenario is the simplest among the three. Nevertheless, \textsf{pH-MARL} is still the best because it better learns the collision avoidance constraint.

\begin{table}
\renewcommand{\arraystretch}{1.4}
\centering
  \caption{Averaged training convergence cumulative reward for the four ablated methods. In all the cases, $n=4$ robots are used.}
  \begin{tabular}{|c|c|c|c|}
    \hline
    \multirow{2}{*}{Method} &
      \multicolumn{3}{c|}{Mean and std of $\boldsymbol{\Bar}{R}$ over $10$ evaluation episodes} 
    \\
    \cline{2-4}
    & Reverse transport & Sampling & Navigation \\
    \hline
    $\textsf{pH-MARL}$ & $\mathbf{213\pm 21} $ & $\mathbf{161\pm 41}$ & $\mathbf{-53\pm 101}$ \\
    \hline
    $\textsf{MLP}$ & ${64} \pm 38$ & ${73}\pm 25$ & ${-280}\pm 98$ \\
    \hline
    $\textsf{MSA}$ & ${57}\pm 43$ & ${82}\pm 32$ & ${-353}\pm 99$ \\
    \hline
    $\textsf{GSA}$ & ${90}\pm 49$ & ${89}\pm 38$ & ${-204}\pm 87$ \\
    \hline
  \end{tabular}
  \label{tab:converge_R}
\end{table}

To assess the scalability of the different architectures, in Fig.~\ref{fig:scalability_results_ablation} each trained neural network is evaluated with a different number of robots, showing the mean and standard deviation of $\Bar{R}$ over $10$ episodes. The first conclusion is that \textsf{pH-MARL} and \textsf{GSA} scale well with the number of robots, achieving the same cumulative reward per robot for all team sizes. In contrast, \textsf{MSA} and \textsf{MLP} do not scale because their architectures considers directly the global observation and action vectors, so their modules are not ready to process an input vector of different dimension. On the other hand, \textsf{pH-MARL} achieves much better performance than the other networks in all the scenarios and team sizes, thus confirming the importance of using physics-informed priors to ease the learning. The single exception is in the navigation example, where the performance deteriorates when the number of robots increases. 

Fig.~\ref{fig:instances_results_ablation} provides some qualitative examples for the three scenarios and helps to understand some limitations of our approach. Beginning with the navigation scenario, we can see that the size of the arena does not scale with the number of robots, it is always $2\times2$m. Hence, taking into account that the robots have a radius of $15$cm, the space is very small to navigate towards the landmarks without colliding. According to the reward, the robots search for safe motions, so they prefer to avoid collisions rather than moving to their landmarks (Fig.~\ref{fig:instances_results_ablation}, navigation with $8$ robots). 
There are other features that do not scale in the reverse transport and sampling scenarios: size of the box, weight of the box, size of the arena and number of Gaussian modes. Nonetheless, \textsf{pH-MARL} is capable of generalizing, scaling and circumventing these issues and achieve the desired tasks. In the reverse transport scenario, even with $16$ robots that are highly packed inside the box, the robots manage to push the box and avoid erratic movements caused by the low weight of the box. In the sampling scenario, robots learn to spread and coordinate to cover more cells when there are more robots. Interestingly, when there are a lot of robots ($n=9$), some robots move to the corners because they annoy the others by increasing the potential number of collisions.

Summing up, the combination of a port-Hamiltonian formulation and self-attention mechanisms leads to superior performance and the desired scalability with the number of robots, learning control policies that are distributed by design and fully exploit the graph structure of the multi-robot system.

\subsection{Comparative results}\label{subsec:comparison}

We compare the performance of our proposed physics-informed multi-agent reinforcement learning approach with other state-of-the-art multi-agent reinforcement learning approaches {that are not physics-informed}. In particular, we compare with: Multi-Agent Deep Deterministic Policy Gradient (\textsf{MADDPG}) \cite{lowe2017multi}, Mean Field Actor Critic (\textsf{MFAC}) \cite{yang2018mean}, Evolutionary Population Curriculum (\textsf{EPC}) \cite{long2019evolutionary}, Distributed multi-Agent Reinforcement Learning with One-hop Neighbors (\textsf{DARL1N}) \cite{wang2022darl1n} {and Multi-Agent Proximal Policy Optimization (\textsf{MAPPO}) \cite{yu2022surprising}}. The comparison results of the first four methods reported in this paper are extracted from \cite{wang2022darl1n}.

We use the food collection, grassland and adversarial scenarios for the comparison. Following the same procedure in \cite{wang2022darl1n}, for the scenarios with an adversarial team (grassland and adversarial), we first use \textsf{MADDPG} to learn the control policies of both teams. Then, the adversarial team control policy is frozen and the policy of the other team is learned using any of the aforementioned approaches.

\begin{figure*}
    \centering
    \begin{tabular}{ccc}
(d) Food collection
    & 
(e) Grassland
    & 
(f) Adversarial 
    \\
\includegraphics[width=0.31\textwidth]{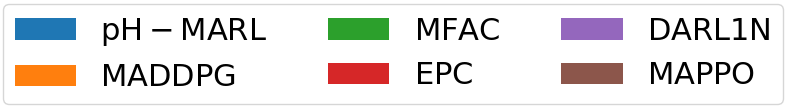}
    & 
\includegraphics[width=0.31\textwidth]{legend_new.png}  
&
\includegraphics[width=0.31\textwidth]{legend_new.png}  
\\
\includegraphics[width=0.31\textwidth, height=0.2\textwidth]{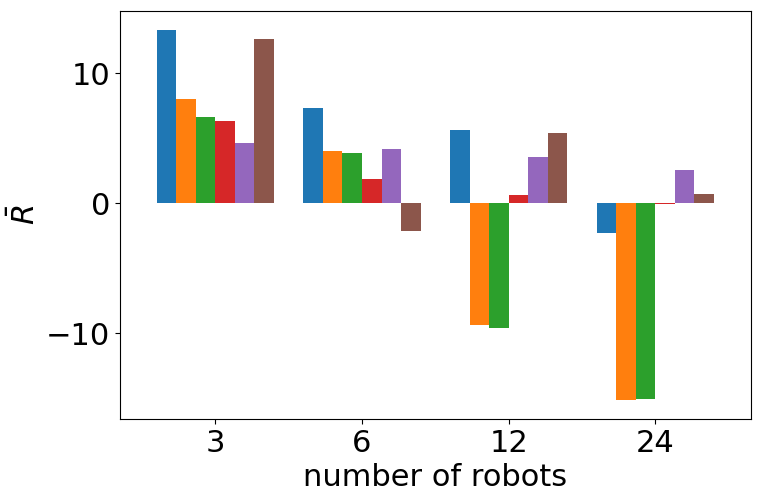}
    & 
\includegraphics[width=0.31\textwidth, height=0.205\textwidth]{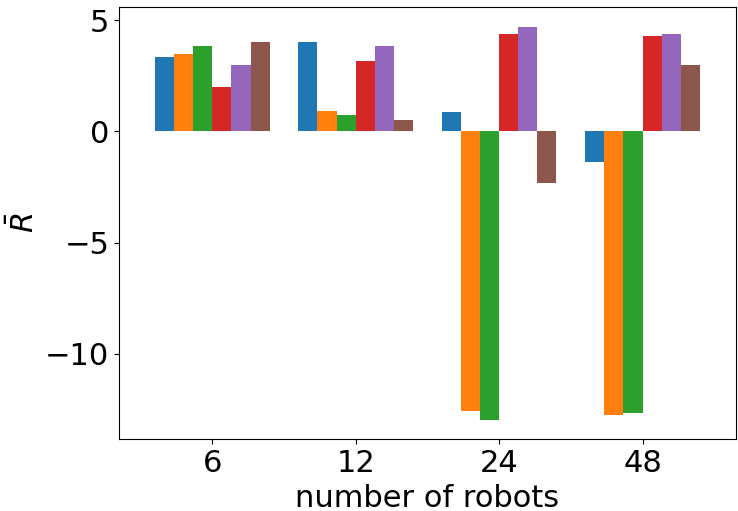}  
&
\includegraphics[width=0.31\textwidth, height=0.2\textwidth]
{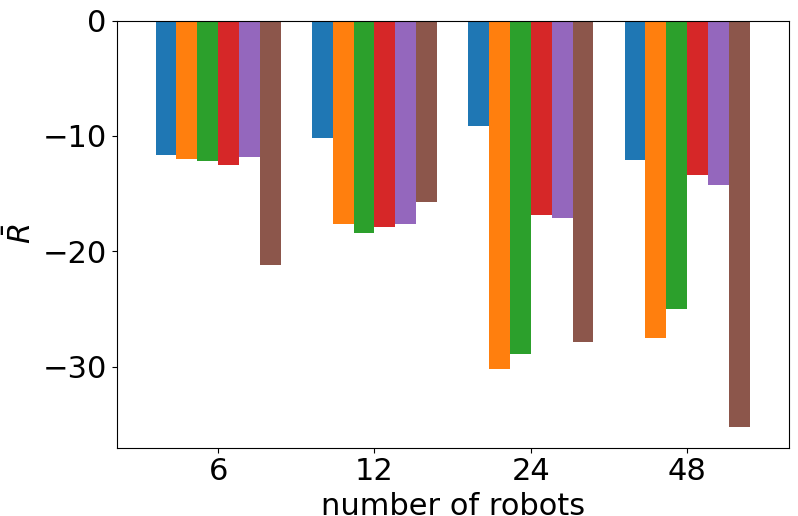}
    \end{tabular}
    \caption{{Comparison of our proposed physics-informed multi-agent reinforcement learning approach with other state-of-the-art approaches. {To measure performance, we use the averaged cumulative reward $\boldsymbol{\Bar}{R} = \frac{1}{n} R = \frac{1}{n}\sum_{\tau=0}^{\tau_{\max} - 1}r(\bfs_\tau,\bfa_\tau)$}. \textsf{pH-MARL} is only trained with $n=4$ robots and deployed with different team sizes, while the state-of-the-art control policies are trained for each specific number of robots. For team sizes similar to the one used in training, our policy outperforms the other approaches. {For team sizes much larger than the training team size, our policy still outperforms or achieves similar results as the state-of-the-art approaches}.}}
    \label{fig:scalability_results_comparison_adhoc}
\end{figure*}

Next, we assess the scalability of our proposed approach in this second series of scenarios. We train the other state-of-the-art approaches for each specific number of robots, ranging from $3$ to $48$ robots. In contrast, we train \textsf{pH-MARL} $n=4$ robots and evaluate the trained control policies with the other number of robots, without further training. 
Fig.~\ref{fig:scalability_results_comparison_adhoc} demonstrates that \textsf{pH-MARL} achieves better or similar results than the other approaches without further training the control policies. When the team size is close to the one used during training ($n=3,6, 12$) \textsf{pH-MARL} surpasses the other methods in the three scenarios, proving the accuracy of combining a physics-informed description of the multi-robot team with self-attention mechanisms. {In this sense, \textsf{MAPPO} in the grassland scenario with $n=6$ robots is the only case that obtains better results than \textsf{pH-MARL}.} When the team size is much greater than the one used during training ($24,48$), \textsf{pH-MARL} outperforms (\textsf{MADDPG}, \textsf{MFAC}) or achieves similar results (\textsf{EPC}, \textsf{DARL1N}, {\textsf{MAPPO}}) as the state-of-the-art. Notably, \textsf{pH-MARL} outperforms all the state-of-the-art methods in the adversarial scenario even when $n=48$, which is significant considering that they are all trained with $n=48$ robots except \textsf{pH-MARL}, which is trained with $n=4$ robots. {Regarding \textsf{MAPPO}, despite its strong performance in some cases (e.g., food collection with $n=3$ robots, grassland with $n=6$ robots or grassland with $n=48$ robots), it generally exhibits poor results when the number of robots is larger than $10$ robots. Besides, \textsf{MAPPO} is unable to learn successful policies that avoid collisions and find the food resources for any of the teams in the adversarial scenario.}


The reason why \textsf{pH-MARL} exhibits a slight drop in performance when $n=24,48$ robots is the same discussed in the ablation experiments. There are some features of the scenarios, like the size of the arena, that do not scale with the number of robots. Robots are not trained to move in configurations that far from those experienced during training. How to design the control policy to be invariant to environmental changes is part of the future work. Qualitative examples can be found in the repository associated to the paper.

{Regarding the computational cost, it is worth noting that, given the same scenario, optimizer and replay buffer hyperparameters, the main differences among methods come from the parameterization of the actor. In this sense, our approach involves greater training times \textit{per sample and robot} compared to other non-physics-informed methods. Specifically, the increase in computational burden comes from the differentiation of the Hamiltonian $\partial H_{\bftheta}(\bfx) / \partial \bfx_j$ at each time step. However, by restricting the space of admissible policies, we have better sample efficiency during training compared to non-physics-informed parameterizations. Therefore, the increase of computational burden per sample and robot is compensated by the sample efficiency of the method, leading to similar overall computational times in robot tasks during training. Besides, as it has been observed in the scalability results, our approach needs a very small number of robots to reproduce or even outperform the other methods when they are trained with larger numbers of robots, saving additional computation. Importantly, the computational cost during deployment scales with the number of neighbors rather than the network size, therefore preserving the desirable properties coming from a distributed multi-robot setting.} {Alternatively, to improve computational efficiency, one could consider a neural network estimator that learns to infer $\partial H_{\bftheta}(\bfx) / \partial \bfx_j$ from $\bfx^i_{\calN^{i,k}}$, which bypasses automatic differentiation of $H_{\bftheta}(\bfx)$ and the three message-passing communication protocol. Avoiding automatic differentiation to reduce inference time is an active area of research \cite{baydin2018automatic}, and we leave its exploration to optimize the training time of our method for future work.}

\begin{figure*}
    \centering
    \begin{tabular}{cccccc}
(a) Reverse transport
    & 
(b) Sampling
    & 
(c) Navigation
\\
\includegraphics[width=0.31\textwidth, height=0.2\textwidth]{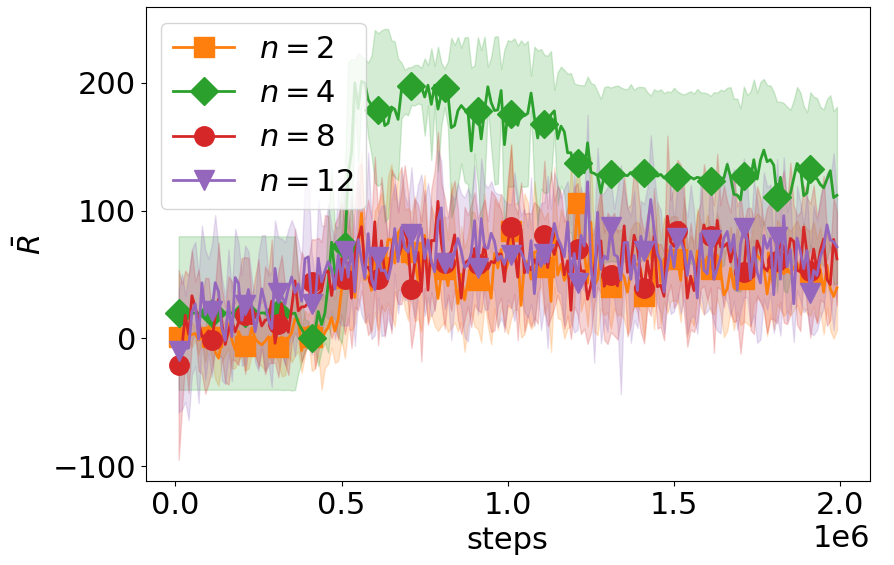}
    & 
\includegraphics[width=0.31\textwidth, height=0.2\textwidth]{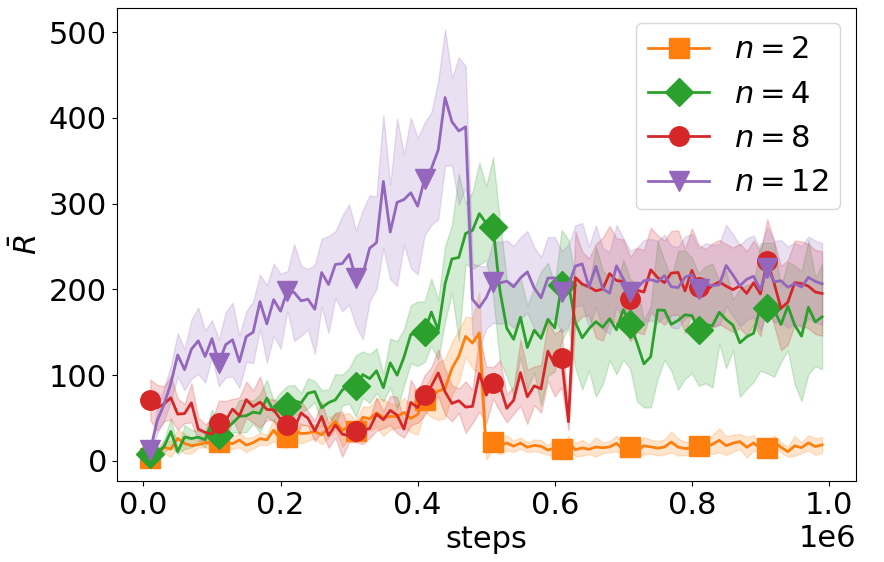}
    & 
\includegraphics[width=0.31\textwidth, height=0.2\textwidth]{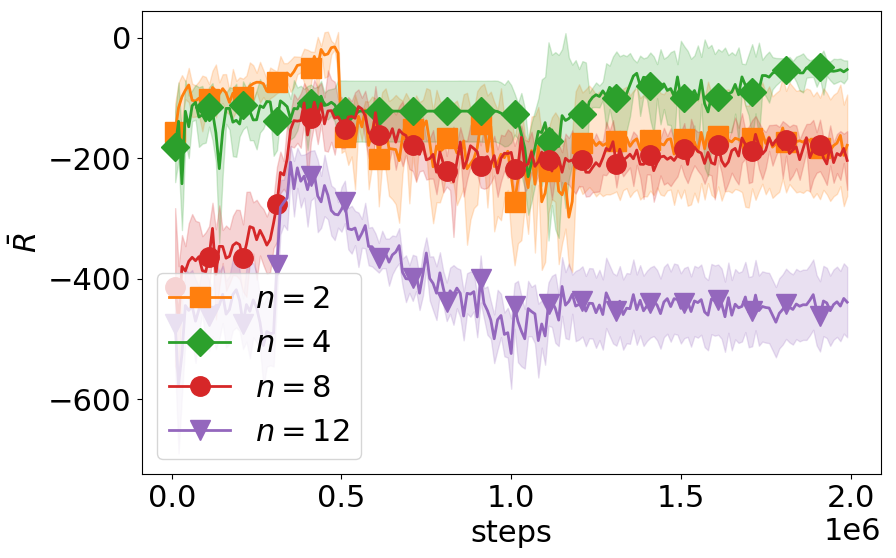}
    \end{tabular}
    \caption{{Sensitivity analysis of \textsf{pH-MARL} with respect to the number of robots at training. To measure performance, we use the averaged cumulative reward $\boldsymbol{\Bar}{R} = \frac{1}{n} R = \frac{1}{n}\sum_{\tau=0}^{\tau_{\max} - 1}r(\bfs_\tau,\bfa_\tau)$. Compared to the nominal value ($n=4$), increasing the number of training robots only leads to greater sample complexity, whereas decreasing the number of training robots does not allow to learn sufficiently varied behaviors to scale with the number of robots.}}
    \label{fig:scalability_training_robots}
\end{figure*}

\subsection{{Sensitivity analysis}}\label{subsec:sensitivity}

{Regarding the evaluation of the training scalability of our approach, we have conducted a series of experiments using the same scenarios of Section \ref{subsec:ablation} and Section \ref{subsec:comparison}. Given those scenarios, we have trained our \textsf{ph-MARL} with an increasing number of robots, namely, $n=\{2, 4, 8, 12\}$ (see Fig. \ref{fig:scalability_training_robots}). The first interesting conclusion is that the increase in number of robots does not necessarily lead to better performance because the interactions among robots always happen at the robot neighborhood level. 
In the reverse transport scenario, more robots inside the cage implies a greater number of robot contacts with the cage and, therefore, a drop in averaged cumulative reward. In the sampling scenario, a greater number of robots leads to an improvement in performance because, for the same time horizon, robots can explore the arena faster and localize the informative regions faster. In the case of the navigation scenario, due to space constraints, an increase in the number of robots during training leads to a significant drop in averaged cumulative reward because the space available to each robot is smaller, so more collisions happen (specially with $n=12$). In all the scenarios, $n=2$ is a too small multi-robot team size to capture the richness of distributed multi-robot interactions, so the training process is unable to learn a successful policy in terms of averaged cumulative reward.}

{Therefore, in general, the optimal number of training robots depends on constraints of the scenario and the task. Nonetheless, as a rule of thumb, increasing the number of training robots towards a large multi-robot team only leads to greater sample complexity, whereas decreasing the number of training robots does not allow to learn sufficiently varied behaviors to scale with the number of robots. Henceforth, we conclude that our approach is best suited for training with a small number of training robots (relative to the task/scenario) but still sufficiently large to capture all possible neighborhood interactions.}

{From a behavioral analysis perspective, the results in Fig. \ref{fig:scalability_training_robots} suggest that the learned distributed multi-robot policies benefit from small neighborhoods, irrespective of the network size. In scenarios with a fixed arena size, an increase in the number of robots indirectly leads to an increase in neighborhood sizes, with a subsequent drop in averaged cumulative reward both in training and evaluation (see Fig. \ref{fig:scalability_results_comparison_adhoc}). The appropriate number of training robots depends on the scenario, e.g., in the reverse transport scenario the best performance is achieved with $n=4$  while in the sampling scenario this is obtained with $n=8$. In any case, as concluded above, our approach requires a sufficiently large team size to capture a variety of local interactions.} {On the other hand, different from other centralized-training decentralized-execution approaches, our approach considers a single set of policy parameters that is invariant to the number of robots in the team due to its distributed structure. Therefore, behavioral analysis to understand individual influence in the overall team reduces to understanding how different neighborhood sizes affect overall performance, which is already given by the sensitivity analysis with respect to the number of training robots for a fixed environment size, and the scalability tests shown in Fig. \ref{fig:scalability_results_ablation} and Fig.~\ref{fig:scalability_results_comparison_adhoc}.}

\subsection{{Validation on a realistic robot platform}}\label{subsec:realistic}

\begin{table}
\renewcommand{\arraystretch}{1.4}
\setlength{\tabcolsep}{2pt}
\centering
  \caption{{Physical properties of the joints of the MuJoCo \textsf{Half Cheetah}.}}
  \begin{tabular}{|c|c|c|c|c|}
    \hline
    Joint & Type & Range [rad] & Damping [Ns/m] & Stiffness [N/m] \\
    \hline
    Back thigh & hinge & $[-0.52, 1.05]$ & $6$ & $240$\\
    \hline
    Back shin & hinge & $[-0.785, 0.785]$ & $4.5$ & $180$ \\
    \hline
    Back foot & hinge & $[-0.4, 0.785]$ & $3$ & $120$ \\
    \hline
    Front thigh & hinge & $[-1, 0.7]$ & $4.5$ & $180$ \\
    \hline
    Front shin & hinge & $[-1.2, 0.87]$ & $3$ & $120$ \\
    \hline
    Front foot & hinge & $[-0.5, 0.5]$ & $1.5$ & $60$ \\
    \hline
  \end{tabular}
  \label{tab:chetaah}
\end{table}

{We further validate \textsf{pH-MARL} in a realistic robot setting using MuJoCo \cite{todorov2012mujoco}. MuJoCo is a general purpose physics engine that replicates multi-joint dynamics with contact, allowing for fast yet realistic physical behaviors. Specifically, we use the Multi-agent MuJoCo benchmark \cite{peng2021facmac}, where the task is to learn coordination policies for multi-joint robots that make them walk. Our approach was evaluated using the \textsf{6x1-Half Cheetah} environment (see Fig.~\ref{fig:overall_scenarios}e). Since our fundamental contribution is the physics-informed distributed policy parameterization, we integrated our policy design in a Trust-Region Policy Optimization (TRPO) reinforcement learning algorithm \cite{schulman2015trust}. This demonstrates the flexibility of the physics-informed policy parameterization in the sense that it can be integrated in different reinforcement learning algorithms.
The benchmark provides results from state-of-the-art multi-agent reinforcement learning approaches for comparison, including Heterogeneous Agent Trust Region Policy Optimization (\textsf{HATRPO}) and Heterogeneous-Agent Proximal Policy Optimization (\textsf{HAPPO}) \cite{kuba2021trust}, \textsf{MAPPO}, \textsf{IPPO} \cite{de2020independent}, and \textsf{MADDPG}. Importantly, in all these state-of-the-art methods, the policy of each agent/joint depends on the global observation vector, whereas our policy only depends on the neighboring information, enforcing a ring topology.} 

\begin{figure}
    \centering
    \begin{tabular}{c}
    \includegraphics[width=0.9\linewidth]{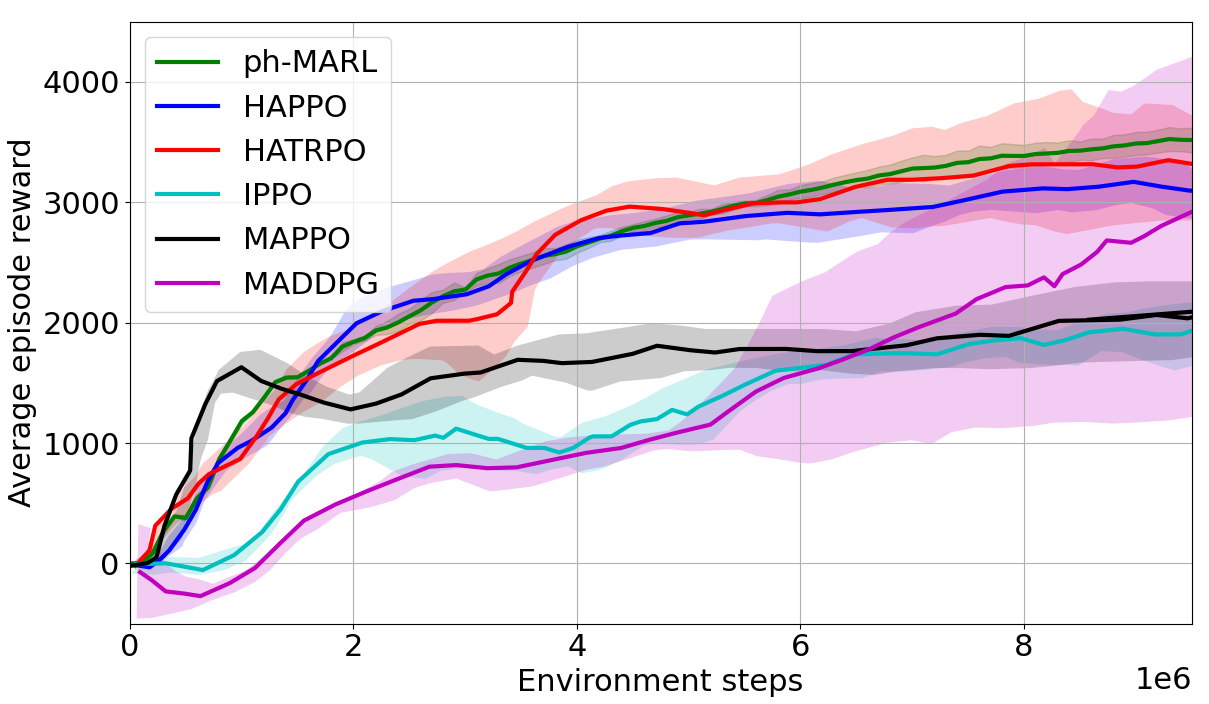} 
    \end{tabular} 
    \caption{{Average episode reward obtained by \textsf{ph-MARL} and other MARL approaches in the \textsf{6x1-Half Cheetah} environment. }}
    \label{fig:mujoco}
\end{figure}

{The \textsf{6x1-Half Cheetah} model is composed of $n=6$ joints, corresponding to: \{back thigh, back shin, back foot, front thigh, front shin, front foot\}. The input to each joint is a torque with [Nm] units. The local observation of each agent is the angle, angular velocity, linear velocity, angular acceleration and linear acceleration of the joint; and the (x,z)-position, angular velocity and (x,z)-linear velocity of front tip. The physical properties of the joints follow those of the original MuJoCo \textsf{Half Cheetah} \cite{todorov2012mujoco}, detailed in Table \ref{tab:chetaah}. Importantly, physical interactions like friction of the joints with the armature of the main body are computed numerically from the stiffness and damping of the joint. Stiffness is modeled as a spring with equilibrium at the nominal position of the joint, whereas the damping is modeled as a force linear in velocity.}

\begin{figure*}
    \centering
\includegraphics[width=\linewidth]{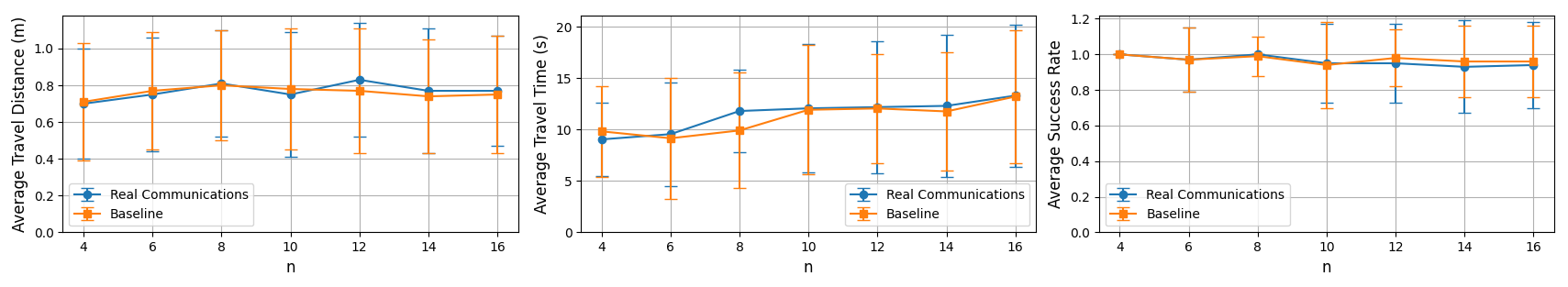}
    \caption{{Quantitative results of the \textsf{pH-MARL} policy for multi-robot navigation in the Robotarium simulator. From left to right, the following metrics are reported for different numbers of robots and communication quality (baseline means perfect communications): travel distance, travel time and success rate. All the metrics are the average over $10$ random runs.}}
    \label{fig:perf_vs_dist}
\end{figure*}

\begin{figure*}
    \centering
    \begin{tabular}{cccc}
(a) $n=4$
    & 
(b) $n=8$
    & 
(c) $n=12$
    &
(d) $n=16$
\\
\includegraphics[width=0.23\linewidth]{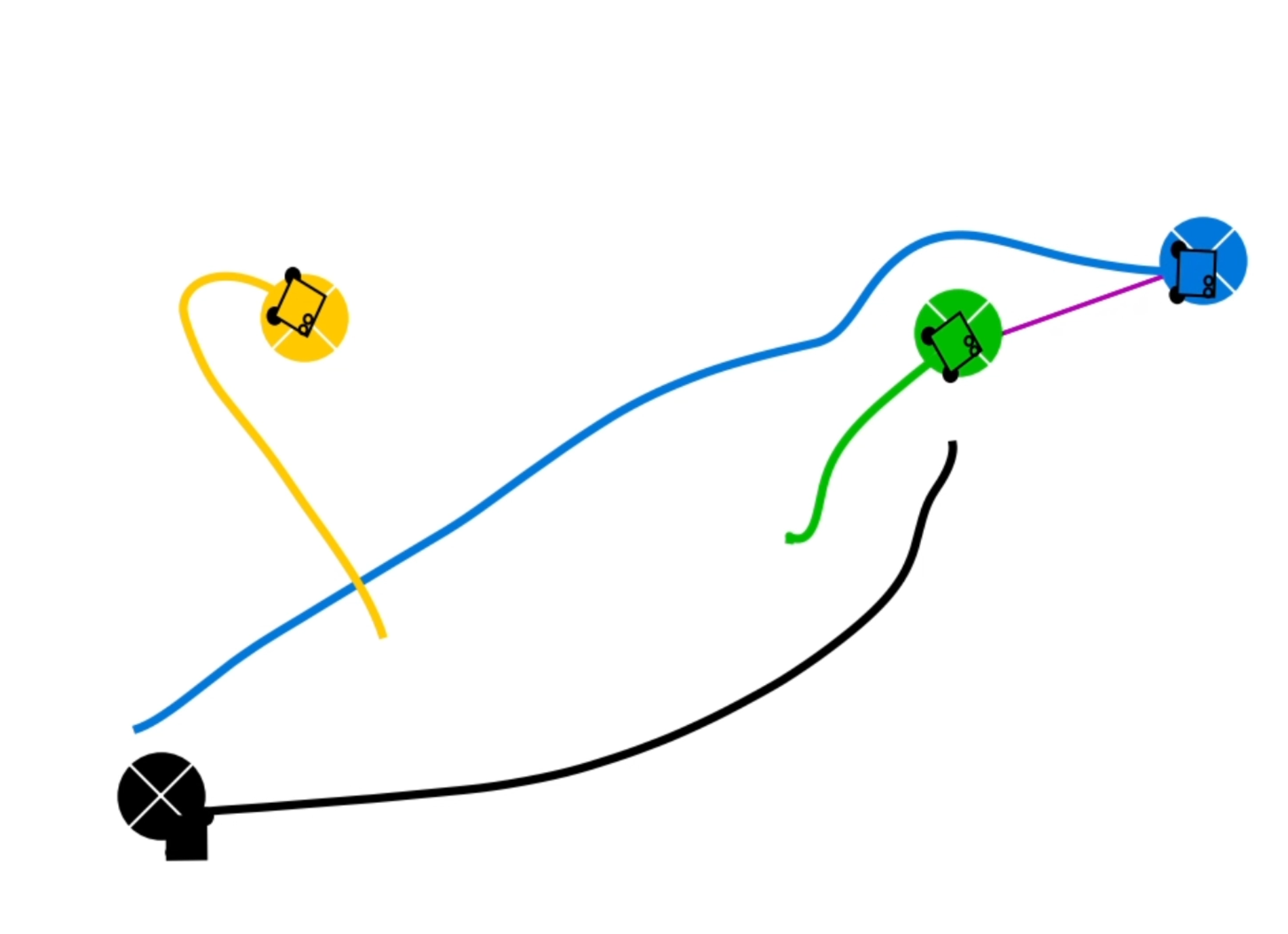}
    & 
\includegraphics[width=0.23\linewidth]{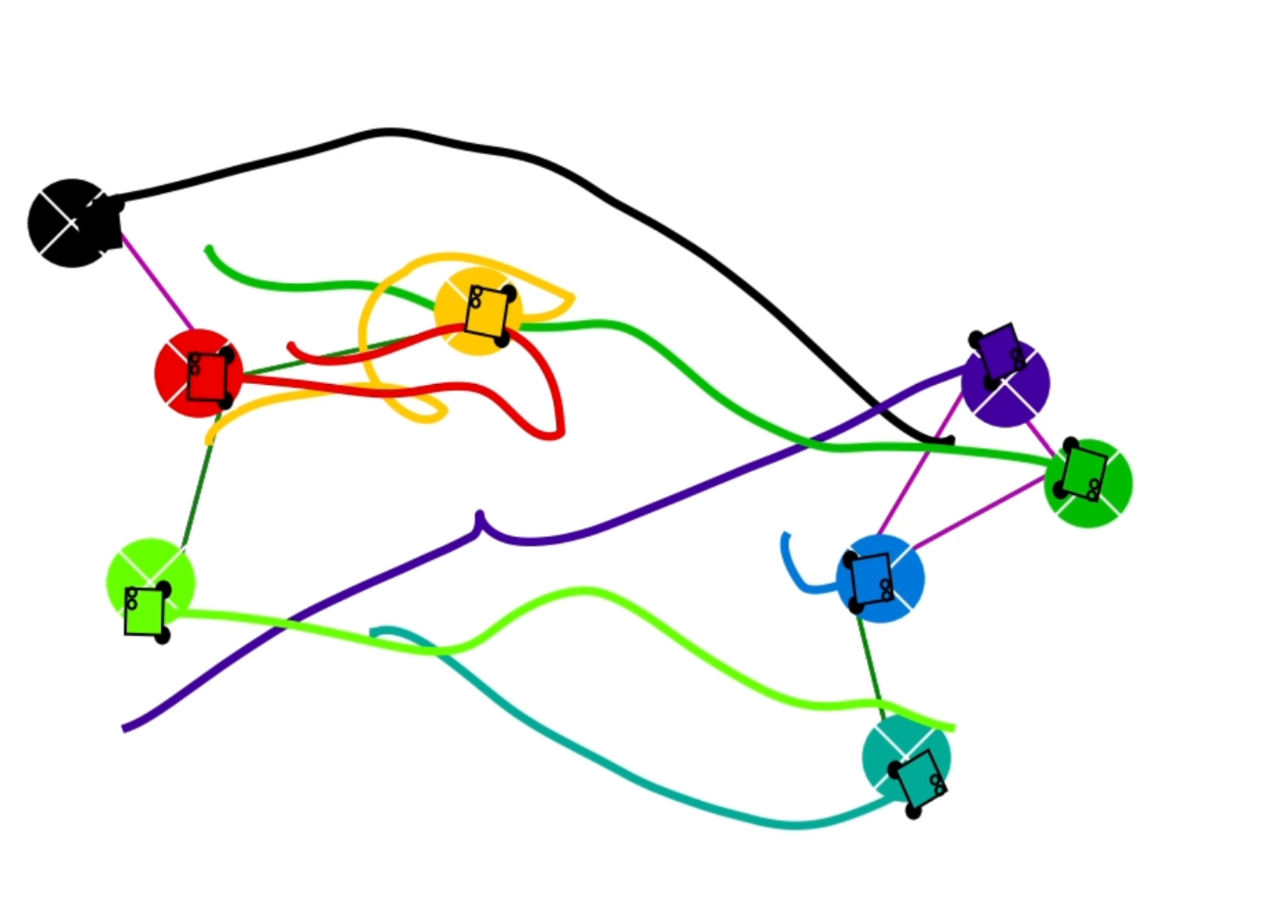}
    & 
\includegraphics[width=0.23\linewidth]{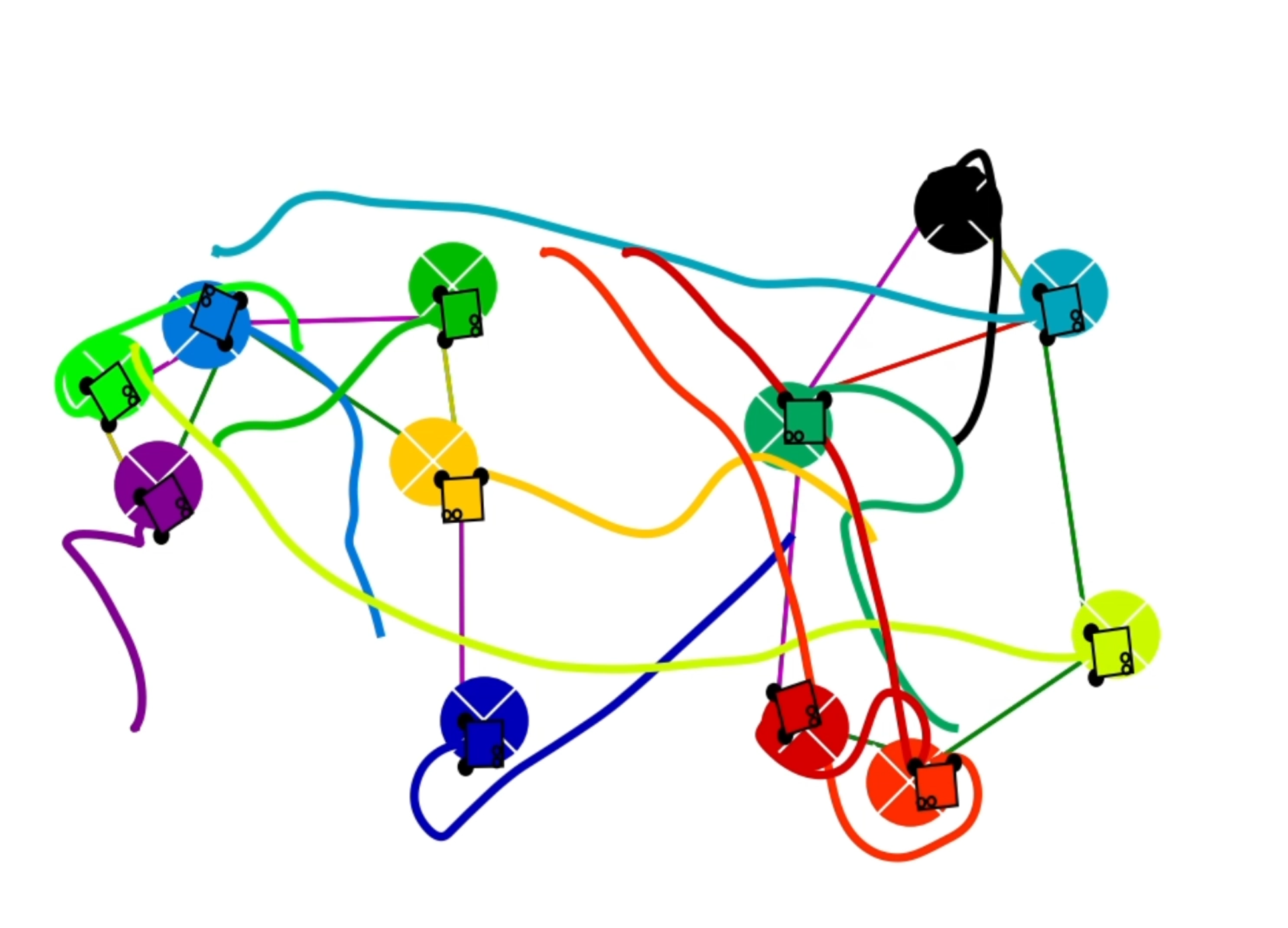}
    & 
\includegraphics[width=0.23\linewidth]{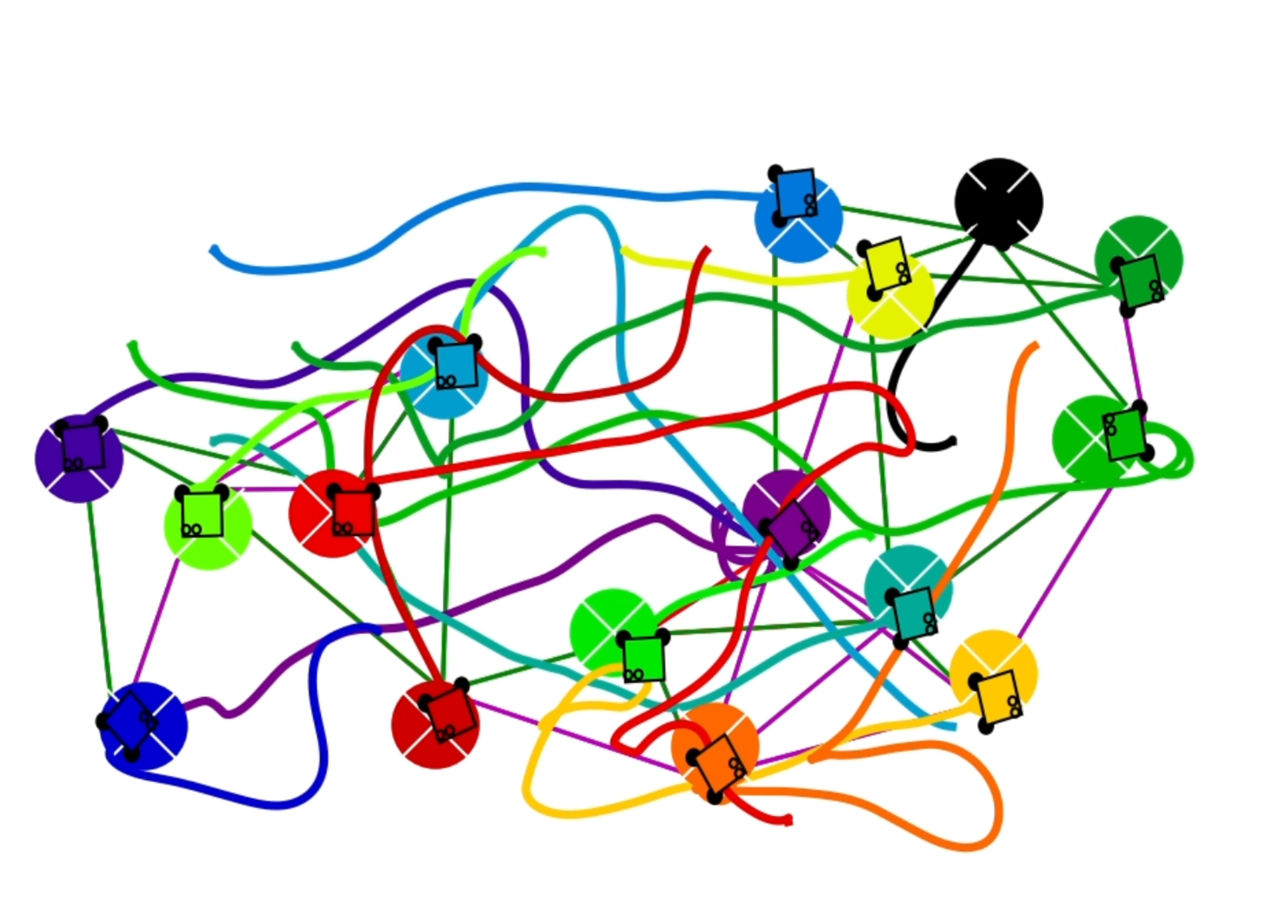}
    \end{tabular}
    \caption{{Qualitative results of the \textsf{pH-MARL} policy for multi-robot navigation in the Robotarium simulator. In (a), communication is perfect, whereas (b), (c) and (d) operate under imperfect communication. In the imperfect communication cases, magenta links denote delays, red links denote packet loss, yellow links denote disturbances, and green links denote unperturbed links.}}
    \label{fig:scalability_robotarium_sim}
\end{figure*}

{As it can be observed in Fig. \ref{fig:mujoco}, our method, although less informed than the others due to the networked constraints, improves upon \textsf{IPPO}, \textsf{MADDPG} and \textsf{MAPPO} in average episode reward, and obtains a similar final average episode reward to \textsf{HAPPO} and \textsf{HATRPO}. Besides, by fully exploiting the physical dynamics of the robots, our method presents better sample efficiency than \textsf{MADDPG}, \textsf{IPPO} and \textsf{MAPPO}, and similar sample efficiency than \textsf{HAPPO} and \textsf{HATRPO}. In this sense, our method leverages the fact that the acceleration of the joints is available, so in the open-loop feed-forward term $\dot{\mathbf{x}}$ is known without explicit knowledge on the values of $\mathbf{J}(\bfx)$, $\mathbf{R}(\bfx)$ and $\partial {H}(\bfx)/\partial\bfx$.
This experiment shows that our approach is a promising method for multi-joint dynamical robots. {The sample efficiency achieved by distributing the policy according to the imposed ring topology is reduced when the connectivity of the network increases. The ring topology restricts the neighborhood of joint $i$ to $\{i-1 \mod 6, i, i+1 \mod 6\}$, which implies a very sparse graph. This serves as a prior for how the other joints affect joint $i$ that restricts the class of policies that can be learned, leading to improved sample efficiency. In the extreme case where the topology is a fully connected graph, our approach would recover the centralized topology of the benchmarked methods. The performance in terms of average episode reward would, therefore, be at least the same, since more information is provided to the joints; however, the sample efficiency would be hampered by computing $6$ centralized policies, one per joint.}


\subsection{{Real multi-robot experiments}}\label{subsec:robotarium}

{Finally, we validate the sim-to-real transfer capabilities of \textsf{pH-MARL} using the Georgia Tech Robotarium \cite{pickem2017robotarium}. The goal of this validation is twofold. First, in all previous sections we assumed perfect communication channels, whereas physical communication in real settings is subject to packet losses, disturbances in the messages and delays, leading to potentially adverse effects on the control policies. Second, we aim to evaluate how \textsf{pH-MARL} policies trained in an ideal simulated environment handle the transfer gap to real multi-robot settings subject to imperfect actuation in the differential-drive robot dynamics of the Robotarium robots. We pose a navigation problem with collision avoidance akin to the ``Navigation'' scenario used for ablations in Section \ref{subsec:ablation}. The only difference is that, to better represent the Robotarium arena, we train over a $3.2 \times 2$m space and increase the physical radius of the agents to $11$cm to fit the size of the Robotarium robots. We train a \textsf{pH-MARL} navigation policy with $n=4$ as in Section \ref{subsec:ablation} in VMAS.}

{After that, we integrate the trained policy in the Robotarium simulator, designed to be as similar as possible to the Robotarium arena to guarantee that policies submitted to the real Robotarium platform comply with all requirements. Two settings are evaluated. The first one evaluates the \textsf{pH-MARL} policy with perfect communications with $r_{comm} = 0.75$m. The second evaluates the \textsf{pH-MARL} policy with realistic communication, as follows.
\begin{itemize}
    \item \textbf{Packet losses}: An existing communication link has probability $0.1$ of being removed from the graph.
    
    \item \textbf{Communication disturbances}: Messages sent through a communication channel have probability $0.1$ of being distorted by an additive zero-mean Gaussian noise with variance $0.05\mathbf{I}$.
    
    \item \textbf{Delays}: Messages sent through a communication channel have probability $0.1$ of suffer a delay uniformly distributed in the range $[1, 10]$ sampling times (sampling time is $0.033$s).
\end{itemize}}
{Fig.~\ref{fig:perf_vs_dist} compares the travel distance, travel time and success rate for the \textsf{pH-MARL} policy under perfect and imperfect communication for different number of robots, averaged over $10$ random runs each. The \textsf{pH-MARL} policy is able to scale to teams up to $\times4$ larger than the one used for training, achieving constant travel distances and almost constant travel times with respect to $n$. The success rates are always above $95\%$, proving that the \textsf{pH-MARL} policy consistently transfers to different robot dynamics (training is done with a damped second-order integrator, whereas the Robotarium robots follow first-order differential-drive dynamics) and environment. Importantly, the \textsf{pH-MARL} policy is robust against imperfect communication, achieving the same results obtained with perfect communication. Qualitatively speaking, Fig. \ref{fig:scalability_robotarium_sim} shows that the robots achieve smooth trajectories and they reach their goals avoiding collisions.}

{Once we verified that the \textsf{pH-MARL} policy is scalable and robust against communication disturbances in the Robotarium simulator, we evaluated it in the real Robotarium arena under imperfect communication. All the parameters are the same, and every configuration of number of robots is averaged over $3$ random runs. As show in Fig. \ref{fig:robotarium_metrics}, the travel distance and travel time are consistent across number of robots, even though the space in the arena becomes increasingly tight with fleets of $12$ and $16$ robots. The case of $n=4$ is much easier than the others, and consequently the travel distance and time are smaller than in the other configurations. The success rate is always $100\%$, allowing us to conclude that our \textsf{pH-MARL} policy supports zero-shot sim-to-real transfer. This is reinforced by the qualitative results in Fig. \ref{fig:scalability_robotarium_real}, where robots follow smooth paths and avoid collisions akin to what the \textsf{pH-MARL} policy achieves in the Robotarium simulator.}

\begin{figure*}
    \centering
\includegraphics[width=\linewidth]{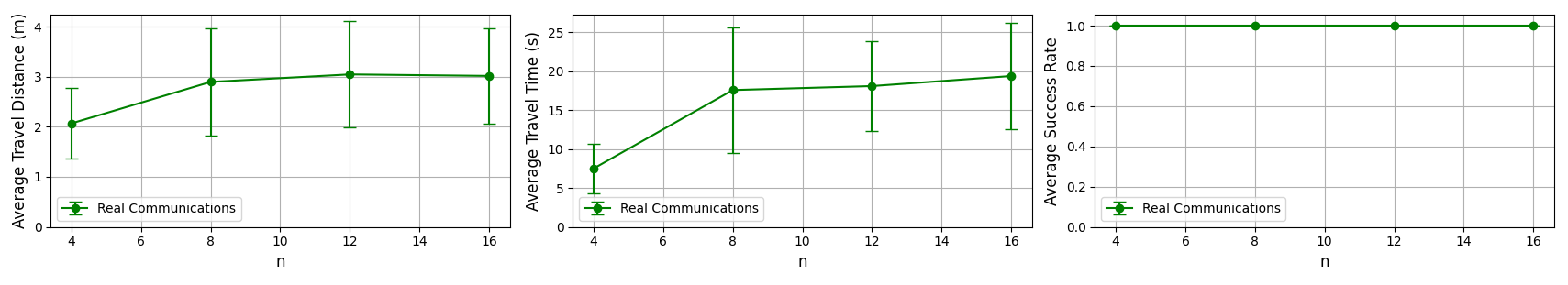}
    \caption{{Quantitative results of the \textsf{pH-MARL} policy for multi-robot navigation in the Robotarium arena under imperfect communication. From left to right: travel distance, travel time and success rate. All the metrics are the average over $3$ random runs.}}
    \label{fig:robotarium_metrics}
\end{figure*}

\begin{figure}
    \centering
    \begin{tabular}{cc}
(a) $n=4$
    & 
(b) $n=8$
\\
\includegraphics[width=0.43\linewidth]{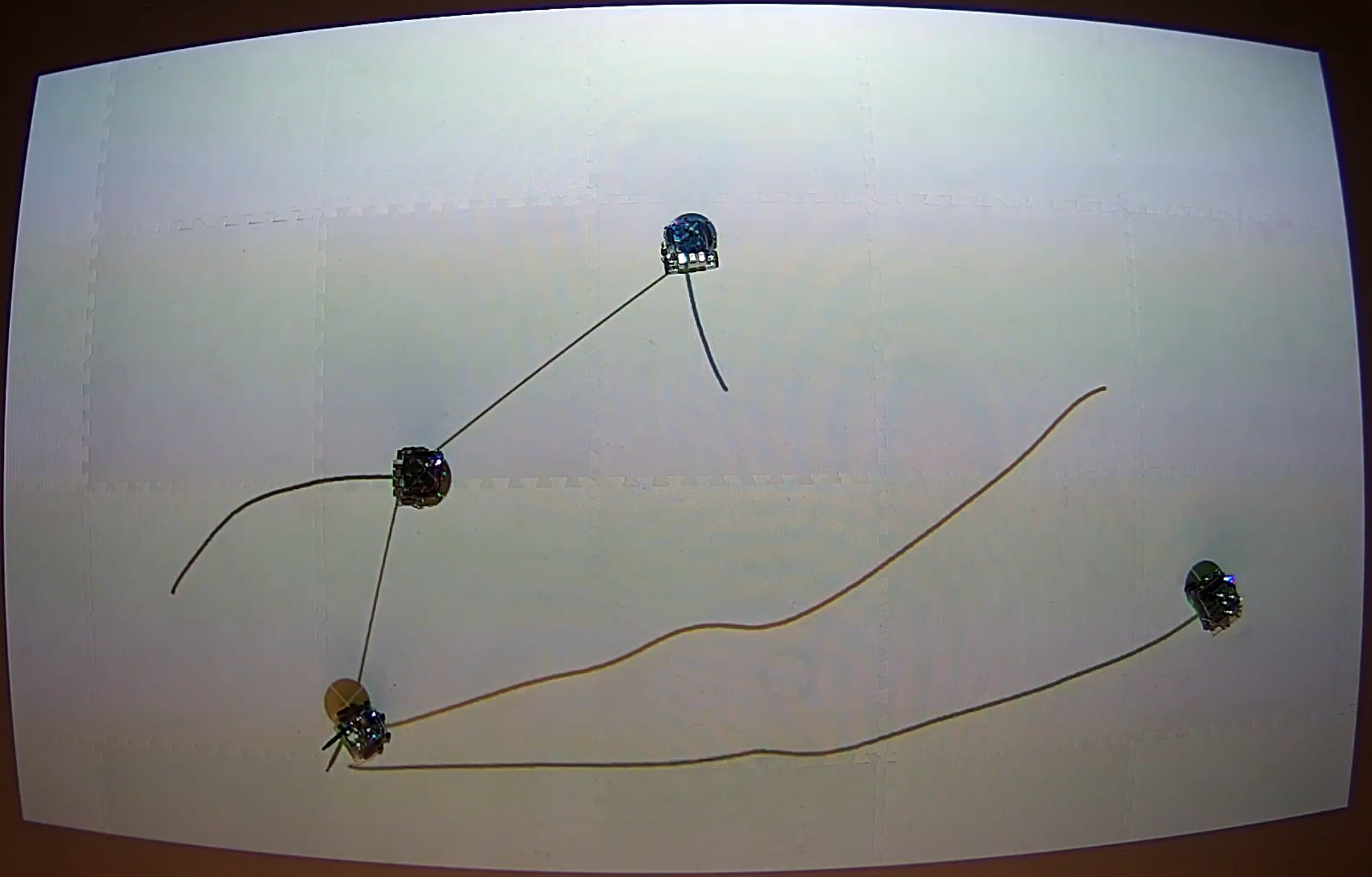}
    & 
\includegraphics[width=0.43\linewidth]{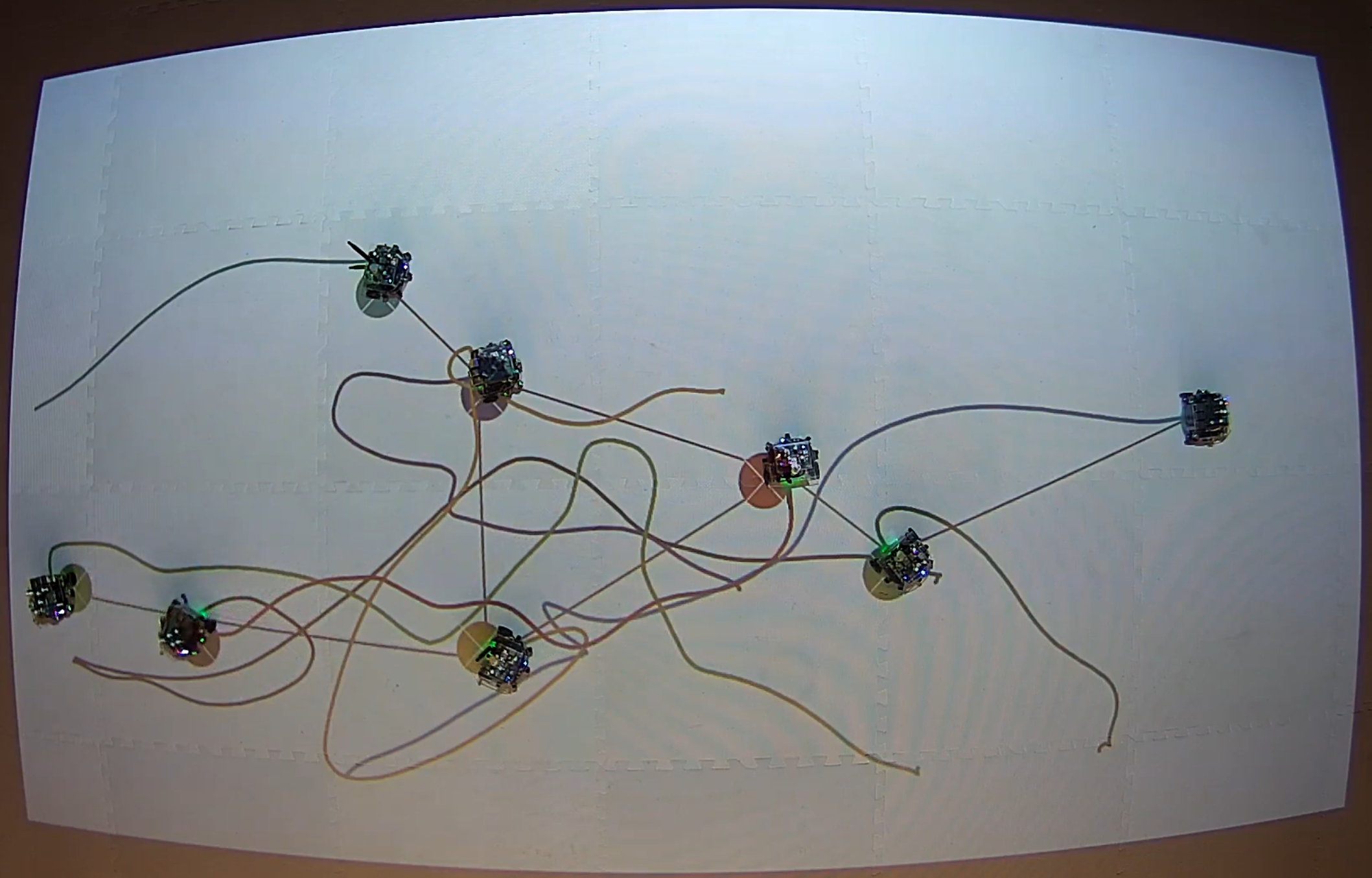}
\\
(c) $n=12$
    &
(d) $n=16$
\\
\includegraphics[width=0.43\linewidth]{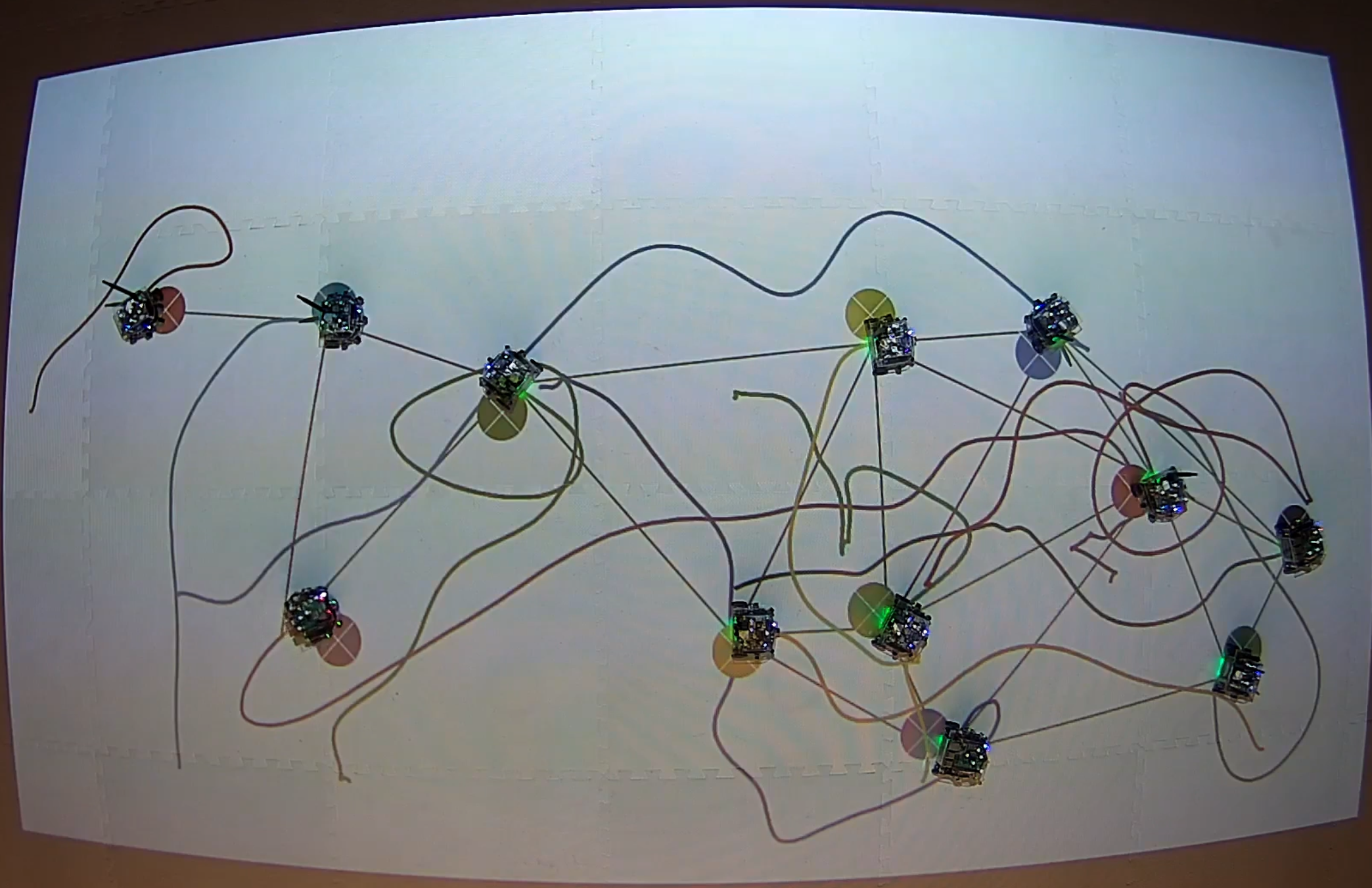}
    & 
\includegraphics[width=0.43\linewidth]{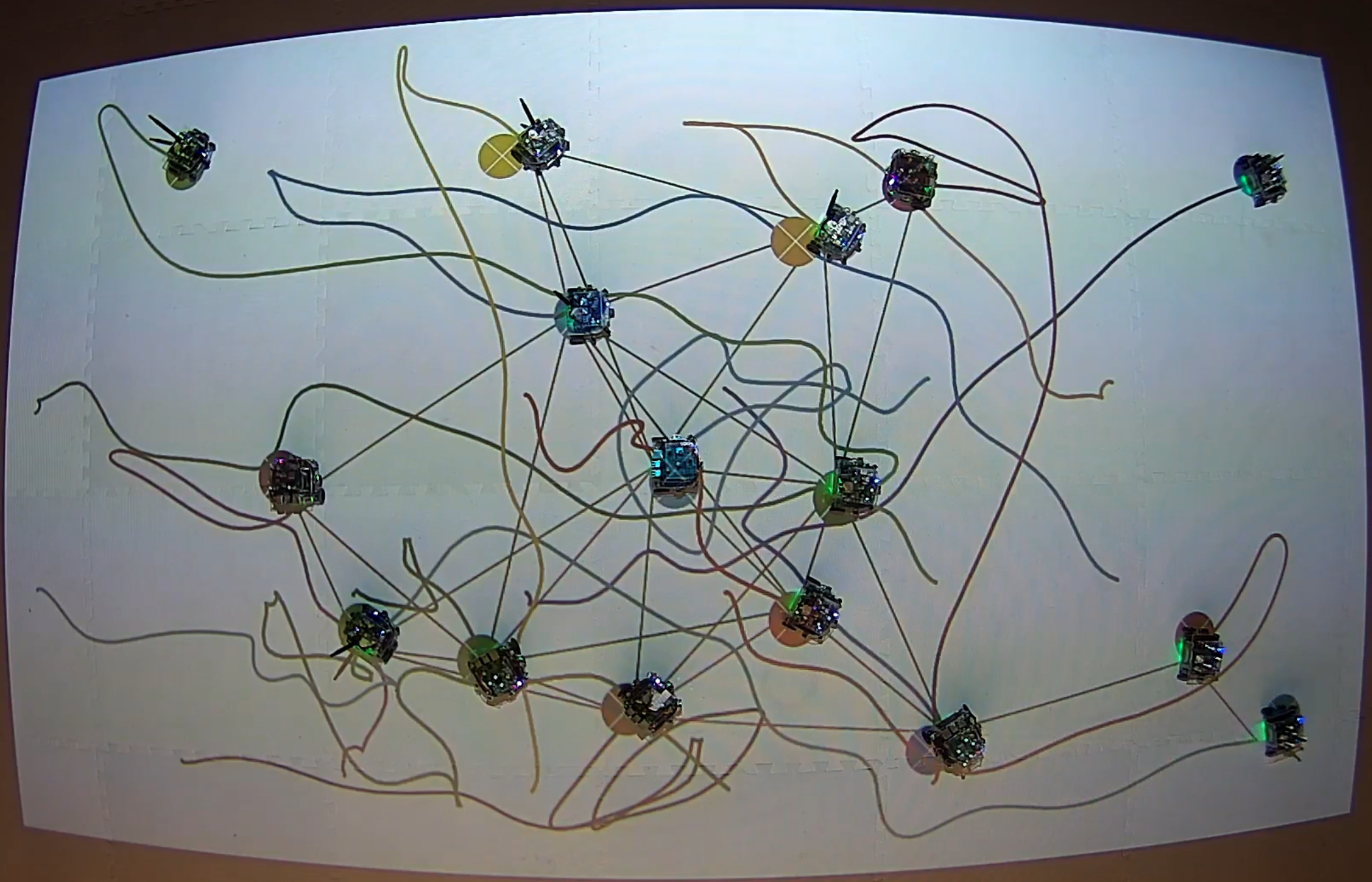}
    \end{tabular}
    \caption{{Qualitative results of the \textsf{pH-MARL} policy for multi-robot navigation in the Robotarium arena under imperfect communication. After training a navigation policy with $n=4$ robots in the VMAS simulator, we directly deployed it in teams of $n=\{4, 8, 12, 16\}$ robots. The figures display coloured trajectories and circles for each robot and associated goal. The robots achieve safe navigation with collision avoidance even in tight spaces with teams $4\times$ larger than those seen during training despite the imperfect communications.}}
    \label{fig:scalability_robotarium_real}
\end{figure}


\section{Conclusions}\label{sec:conclusion}
We proposed a novel reinforcement learning formulation where the single agent is the multi-robot graph. On the one hand, this allowed to explicitly consider the potential networked interactions among agents, going beyond the classical assumptions of independent execution found in other multi-agent reinforcement learning approaches. On the other hand, this allowed to avoid non-stationarity issues found during training in other multi-robot learning setting. In particular, we designed a soft actor-critic algorithm to manage the 
networked and stochastic nature of distributed multi-robot policies, learning simultaneously the homogeneous distributed control policy for all the robots while respecting the policy $k$-hop factorization and the correlation among robots. Besides, the method exploits the collective knowledge of a centralized critic during training. To achieve the learning of scalable and distributed control policies by design, we proposed an interconnection and damping passivity-based control policy based on a port-Hamiltonian description of the multi-robot dynamics that preserves energy conservation laws and individual robot dynamics. To parameterized the controller, we proposed a set of self-attention-based neural networks that respects the desired distributed structure of the control policy and handles the time-varying available information at each robot.

We conducted ablation studies and simulations in comparison to state-of-the-art multi-agent reinforcement learning approaches in seven scenarios, covering a wide variety of cooperative and competitive behaviors such as collision avoidance, navigation, transport, evasion and monitoring, {and including a validation experiment in a realistic robotic platform}. In all the cases, our proposed approach  exhibits superior performance in terms of cumulative reward per robot and scalability. Our approach, without further training, scales and achieves the same performance of other methods that are trained, ad hoc, with a specific number of robots. {Extensive validation in a real multi-robot navigation setting demonstrated that policies trained in an ideal simulator exhibit great zero-shot sim-to-real transfer capabilities, even under imperfect communication.} Nevertheless, there is still room for improvement in terms of generalization with respect to environmental conditions such as stage size. These results lead to the following conclusion: the combination of physics-informed neural networks and reinforcement learning techniques is a promising research line to address multi-robot problems. 

{A important future direction is to utilize the physics-informed port-Hamiltonian parameterization of the control policies to provide performance or safety guarantees. For instance, passivity theory or control barrier functions can be employed to ensure convergence properties of the control policy or design safety modules on top of the control policy that provide collision avoidance guarantees. It also remains to be explored how to bypass the three-message passing scheme to compute $\partial H_{\bftheta}(\mathbf{x})/\partial\mathbf{x}_j$ and rely only on single-message broadcasting protocols or perception cues since it is currently a computation and implementation bottleneck.}


\appendices
\section{Network Parameters}\label{sec:parameters}
The self-attention-based control policy in \textsf{ph-MARL} is parameterized as follows:
\begin{itemize}
    \item $[\mathbf{R}_{\bftheta}]_{ij}$: $W = 3$, $h_w=[n_x, 16, 8]$, $r_w = [8, 16, 8]$, $d_w = [16, 8, 16]$; functions $\beta = \mathsf{sigmoid}$, $\chi=\psi=\mathsf{swish}$~\cite{ramachandran2017searching}. 
    \item $[\mathbf{J}_{\bftheta}]_{ij}$: $W = 3$, $h_w=[n_x, 16, 8]$, $r_w = [8, 16, 8]$, $d_w = [16, 8, 16]$; functions $\beta = \mathsf{sigmoid}$, $\chi=\psi=\mathsf{swish}$~\cite{ramachandran2017searching}.
    \item $H^{i}_{\bftheta}$: $W = 3$ layers, $h_w=[n_x, 16, 8]$, $r_w = [16, 8, 8]$, $d_w = [16, 8, 25]$; functions $\beta = \mathsf{sigmoid}$, $\chi=\psi=\mathsf{swish}$~\cite{ramachandran2017searching}.
\end{itemize}
For the variance network we use the same architecture of $[\mathbf{R}_{\bftheta}]_{ij}$ but with $h_1 = n_x + n_u$ and $d_W = 2$. The other networks used in the ablation studies are as follows: 
\begin{itemize}
    \item $\mathsf{MLP}$: the policy is a single multi-layer perceptron of size $[n \times n_x, n \times n_u, n \times n_u]$ with swish hidden activation function and linear output. For the variance network we use the same architecture but with size $[n (n_x + n_u), n \times n_u, n \times n_u]$. 
    \item $\mathsf{MSA}$: the policy is a single multi-layer perceptron of size $[n \times n_x, n \times n_x]$ with linear output followed by a self-attention layer where query, key and values are directly the feature vector from the multi-layer perceptron, and an additional multi-layer perceptron of size $[n \times n_x, n \times n_u]$ with linear output. For the variance network we use the same architecture but with size $[n (n_x + n_u), n \times n_u]$ for the first multi-layer perceptron. 
    
    \item $\mathsf{GSA}$: the policy is the same one used to predict $[\bfR_\bftheta]_ij$ but with $d_W = 2$. For the variance network we use the same architecture but $h_1=n_x + n_u$.  
\end{itemize}
The $Q_{\bfPi_\bftheta}(\bfs_t, \bfa_t)$ function is always parameterized as a multi-layer perceptron with layers of size $[n(n_x + n_u), 2n(n_x + n_u), n(n_x + n_u), (n_x + n_u), 1]$ with swish \cite{ramachandran2017searching} activation functions except the last layer, that is linear.

The food collection, grassland, and adversarial scenarios use a neural network to pre-process the observation vector to move from a time-varying observation size to a fixed state size compatible with the control policies. To do so, the state is a concatenation of the position (2-dimensional vector), velocity (2-dimensional vector), aliveness (boolean quantity), closest goal relative distance (2-dimensional vector) and a 2-dimensional feature vector provided by a neural network. The neural network is composed by a multi-layer perceptron with no hidden layers of size $[n_x - 2, 2]$ and swish activation function, a self-attention layer of size $h_w = r_w=d_w=2$, and another multi-layer perceptron with no hidden layers of size $[2, 2]$ and linear output.

\section{Soft Actor-Critic Hyperparameters}\label{sec:sac_parameters}
The following table details the parameterization of the soft actor-critic algorithm for the different scenarios.
\begin{longtblr}[
  caption = {Soft actor-critic hyperparameters.},
  entry = {Short Caption},
  label = {tab:sac_parameters}
]{colsep  = 2pt, colspec = {||c|c|c||}}
\hline
\SetCell{c=1}{\textbf{Parameter}}        & \SetCell{c=1}{\textbf{Scenario}}       & \SetCell{c=1}{\textbf{Value}} 
\\ 
\hline
\SetCell{c=1}{\multirow{1}{*}{optimizer}} 
& \SetCell{c=1}{all}   
& \SetCell{c=1} {Adam~\cite{kingma2014adam}} 
\\ 
\hline
\SetCell{c=1}{\multirow{3}{*}{$r_{comm}$}} 
& \SetCell{c=1}{\pbox{4.5cm}{reverse transport, navigation}}   
& \SetCell{c=1}{$0.45$m} 
\\ 
\cline{2-3} 
\SetCell{c=1}{}                  
& \SetCell{c=1}{sampling}   
& \SetCell{c=1}{$0.75$m}
\\ 
\cline{2-3} \SetCell{c=1}{}             & \SetCell{c=1}{\pbox{4.5cm}{food collection, grassland, adversarial}}   
&  \SetCell{c=1}{$0.15$m}
\\ 
\hline
\SetCell{c=1}{\multirow{2}{*}{$n$ training}} 
& \SetCell{c=1}{\pbox{4.5cm}{reverse transport, sampling, navigation}}   
& \SetCell{c=1}{$4$} 
\\ 
\cline{2-3} \SetCell{c=1}{}             & \SetCell{c=1}{\pbox{4.5cm}{food collection, grassland, adversarial}}   &  \SetCell{c=1}{$8$} \\ \hline
\SetCell{c=1}{\multirow{1}{*}{\pbox{1.5cm}{\# parallel environments}}} & \SetCell{c=1}{\multirow{1}{*}{all}}   & \SetCell{c=1}{\multirow{1}{*}{$96$}} 
\\
\hline
\SetCell{c=1}{\multirow{1}{*}{\pbox{1.5cm}{shared $r$}}} & \SetCell{c=1}{\multirow{1}{*}{all}}   & \SetCell{c=1}{\multirow{1}{*}{True}} 
\\
\hline
\SetCell{c=1}{\multirow{2}{*}{\pbox{1.5cm}{maximum steps per \\ episode}}} & \SetCell{c=1}{\pbox{4.5cm}{reverse transport, navigation, food collection, grassland, adversarial}}   & \SetCell{c=1}{$400$} \\ \cline{2-3} 
\SetCell{c=1}{}                  & \SetCell{c=1}{sampling}   &  \SetCell{c=1}{$1000$} \\ \hline
\SetCell{c=1}{\multirow{1}{*}{\pbox{1.5cm}{\pbox{1.5cm}{replay buffer size}}}} & \SetCell{c=1}{\multirow{1}{*}{all}}   & \SetCell{c=1}{\multirow{1}{*}{$2\times10^6$}} 
\\ 
\hline
\SetCell{c=1}{\multirow{1}{*}{\pbox{1.5cm}{\pbox{1.5cm}{initial random steps}}}} & \SetCell{c=1}{\multirow{1}{*}{all}}   & \SetCell{c=1}{\multirow{1}{*}{$10^3$}} 
\\ 
\hline
\SetCell{c=1}{\multirow{1}{*}{\pbox{1.5cm}{$\gamma$}}} & \SetCell{c=1}{\multirow{1}{*}{all}}   & \SetCell{c=1}{\multirow{1}{*}{$0.99$}} 
\\ 
\hline
\SetCell{c=1}{\multirow{1}{*}{\pbox{1.5cm}{$\alpha_0$}}} & \SetCell{c=1}{\multirow{1}{*}{all}}   & \SetCell{c=1}{\multirow{1}{*}{$5$}} 
\\ 
\hline
\SetCell{c=1}{\multirow{1}{*}{\pbox{1.5cm}{$\alpha_{\min}$}}} & \SetCell{c=1}{\multirow{1}{*}{all}}   & \SetCell{c=1}{\multirow{1}{*}{$0.1$}} 
\\ 
\hline
\SetCell{c=1}{\multirow{1}{*}{\pbox{1.5cm}{$\alpha_{\max}$}}} & \SetCell{c=1}{\multirow{1}{*}{all}}   & \SetCell{c=1}{\multirow{1}{*}{$10$}} 
\\
\hline
\SetCell{c=1}{\multirow{1}{*}{\pbox{1.5cm}{$\rho$}}} & \SetCell{c=1}{\multirow{1}{*}{all}}   & \SetCell{c=1}{\multirow{1}{*}{$0.005$}} 
\\ 
\hline
\SetCell{c=1}{\multirow{1}{*}{\pbox{1.5cm}{learning rate $\alpha$}}} & \SetCell{c=1}{\multirow{1}{*}{all}}   & \SetCell{c=1}{\multirow{1}{*}{$10^{-5}$}}
\\ 
\hline
\SetCell{c=1}{\multirow{1}{*}{\pbox{1.5cm}{learning rate}}} & \SetCell{c=1}{\pbox{4.5cm}{all}}   & \SetCell{c=1}{$10^{-4}$} 
\\ 
\hline
\SetCell{c=1}{\multirow{1}{*}{\pbox{1.5cm}{\pbox{1.5cm}{batch size}}}} & \SetCell{c=1}{\multirow{1}{*}{all}}   & \SetCell{c=1}{\multirow{1}{*}{$1024$}} 
\\ 
\hline
\SetCell{c=1}{\multirow{2}{*}{\pbox{1.5cm}{\# training steps}}} & \SetCell{c=1}{\pbox{4.5cm}{navigation, reverse transport, sampling, food collection, grassland}}   & \SetCell{c=1}{$2\times10^6$} \\ \cline{2-3} 
\SetCell{c=1}{}                  & \SetCell{c=1}{adversarial}   &  \SetCell{c=1}{$6\times10^5$} \\ \hline
\SetCell{c=1}{\multirow{1}{*}{\pbox{1.5cm}{clip \\ gradients}}} & \SetCell{c=1}{\multirow{1}{*}{all}}   & \SetCell{c=1}{\multirow{1}{*}{False}} 
\\ 
\hline
\SetCell{c=1}{\multirow{1}{*}{\pbox{1.5cm}{reward \\ scaling}}} & \SetCell{c=1}{\multirow{1}{*}{all}}   & \SetCell{c=1}{\multirow{1}{*}{False}} 
\\ 
\hline
\SetCell{c=1}{\multirow{1}{*}{\pbox{1.5cm}{$\sigma_{\min}$}}} & \SetCell{c=1}{\multirow{1}{*}{all}}   & \SetCell{c=1}{\multirow{1}{*}{$e^{-5}$}} 
\\ 
\hline
\SetCell{c=1}{\multirow{1}{*}{\pbox{1.5cm}{$\sigma_{\max}$}}} & \SetCell{c=1}{\multirow{1}{*}{all}}   & \SetCell{c=1}{\multirow{1}{*}{$e^2$}} 
\\ 
\hline
\SetCell{c=1}{\multirow{2}{*}{\pbox{1.5cm}{landmark mass}}} & \SetCell{c=1}{\pbox{4.5cm}{sampling, navigation, food collection, grassland, adversarial}}   & \SetCell{c=1}{default} \\ \cline{2-3} 
\SetCell{c=1}{}                  & \SetCell{c=1}{reverse transport}   &  \SetCell{c=1}{$1$}
\\ 
\hline
\SetCell{c=1}{\multirow{1}{*}{\pbox{1.5cm}{evaluation interval}}} & \SetCell{c=1}{\multirow{1}{*}{all}}   & \SetCell{c=1}{\multirow{1}{*}{$10^4$}} 
\\ 
\hline
\SetCell{c=1}{\multirow{1}{*}{\pbox{1.5cm}{\# evaluation episodes per interval}}} & \SetCell{c=1}{\multirow{1}{*}{all}}   & \SetCell{c=1}{\multirow{1}{*}{$10$}}\\ \hline
\end{longtblr}

\bibliographystyle{IEEEtran}
\bibliography{IEEEabrv,IEEEexample.bib}

\end{document}